\documentclass[11pt]{article}

\usepackage[preprint]{acl}

\usepackage{times}
\usepackage{latexsym}

\usepackage[T1]{fontenc}

\usepackage[utf8]{inputenc}

\usepackage{microtype}

\usepackage{inconsolata}

\usepackage{graphicx}

\usepackage{hyperref}       
\usepackage{url}            
\usepackage{booktabs}       
\usepackage{amsfonts}       
\usepackage{nicefrac}       
\usepackage{microtype}      
\usepackage{xcolor}         
\usepackage{enumitem}
\usepackage{hyperref}
\usepackage{amsmath}
\usepackage{adjustbox}
\usepackage{multirow}
\usepackage{tabularx} 
\usepackage{placeins}
\usepackage{graphicx}
\usepackage{wrapfig}
\usepackage{algorithm}
\usepackage{algpseudocode} 
\usepackage{bbm}
\usepackage[most,breakable,skins]{tcolorbox}
\usepackage{graphicx}
\usepackage{colortbl}
\definecolor{lightgreen}{rgb}{0.88, 1, 0.88}
\usepackage{subcaption}
\usepackage{array}
\usepackage{pifont}
\usepackage{amsmath}
\usepackage{amssymb}
\usepackage{mathtools}
\usepackage{amsthm}
\usepackage{float}

\usepackage{xspace}
\usepackage{xcolor}
\definecolor{ourblue}{HTML}{1F4E79}
\newcommand{\ours}{\textcolor{ourblue}{\textsc{AIRL-S}}\xspace}
\newcommand{\ourpolicy}{\text{Qwen2.5-7B-AIRL-S}\xspace}

\title{AIRL-S: Unifying Reinforcement Learning and Search-Based Test-Time Scaling via Adversarial Inverse Reinforcement Learning}

\author{
\mdseries
Can Jin\textsuperscript{1},
Yang Zhou\textsuperscript{1},
Qixin Zhang\textsuperscript{2},
Hongwu Peng\textsuperscript{3},
Di Zhang\textsuperscript{4},
Zihan Dong\textsuperscript{1},
\\
Marco Pavone\textsuperscript{5},
Ligong Han\textsuperscript{6,7},
Zhang-Wei Hong\textsuperscript{8},
Tong Che\textsuperscript{5\textdagger},
Dimitris N. Metaxas\textsuperscript{1\textdagger}
\\
\textsuperscript{1}Rutgers University \quad
\textsuperscript{2}Nanyang Technological University \quad
\textsuperscript{3}University of Connecticut \\
\textsuperscript{4}Fudan University \quad
\textsuperscript{5}NVIDIA Research \quad
\textsuperscript{6}Red Hat AI Innovation \\
\textsuperscript{7}MIT-IBM Watson AI Lab \quad
\textsuperscript{8}Massachusetts Institute of Technology
\\
\textsuperscript{\textdagger}Equal Advising
}

\begin{document}
\maketitle

\begin{abstract}
Test-time scaling strategies for Large Language Models predominantly rely on either reinforcement learning with sparse outcome rewards or search-based methods guided by static Process Reward Models. However, outcome-based RL often suffers from training instability and sample inefficiency, while static PRMs require expensive step-wise supervision and are susceptible to reward hacking due to distributional shifts. In this paper, we introduce \ours, a unified framework that integrates Adversarial Inverse Reinforcement Learning with Group Relative Policy Optimization. By inferring a dense, step-wise reward model directly from reference trajectories, \ours eliminates the dependency on labeled process data and uses the same learned PRM as both a training signal and a verifier for search-based TTS. Extensive evaluations across eight benchmarks in mathematics, science, and code generation demonstrate that our policy model improves average performance by $\textbf{9\%}$ over the base model, matching GPT-4o. We further analyze how the AIRL and GRPO objectives complement each other and how the learned PRM transfers across generators and search algorithms, establishing a robust and cost-effective methodology for scaling test-time computation in complex reasoning tasks.
\end{abstract}    
\section{Introduction}
\label{sec:intro}
Test-time scaling (TTS) has recently emerged as a primary strategy for enhancing the reasoning capabilities of Large Language Models (LLMs)~\citep{o1,r1,snell2025scaling,s1,zhou2024language,rebase}. To support TTS on complex reasoning benchmarks, researchers have adopted Reinforcement Learning (RL) methods~\citep{o1,r1,deepseekmath2024} and search strategies, such as Monte Carlo Tree Search (MCTS), beam search, and Best-of-N sampling~\citep{zhou2024language,zhang2023planning,llama_berry,snell2025scaling,xie2023self}. Prominent models, including OpenAI's \emph{o}-series~\citep{o1} and DeepSeek-R1~\citep{r1}, demonstrate that large-scale RL training encourages the generation of extended Chains of Thought (CoT) during inference. This training process is typically governed by Outcome Reward Models (ORMs)~\citep{cobbe2021training,r1,yu2023outcome} and Process Reward Models (PRMs)~\citep{li2023making,lightman2023let,uesato2022solving,wang2024math,deepseekmath2024}, which provide the necessary supervisory signals to optimize model performance.

While DeepSeek-R1 demonstrates that strong TTS performance is attainable using only ORMs or rule-based rewards~\citep{r1}, the sparsity of outcome-based signals often compromises training stability and sample efficiency~\citep{deepseekmath2024,lightman2023let,yuan2024free,chan2024dense}. Conversely, search-based TTS methods utilizing PRMs benefit from dense, step-wise guidance. However, training effective static PRMs requires granular supervision, necessitating expensive human annotation or pseudo-labels generated by other LLMs~\citep{wang2024math,lightman2023let,deepseekmath2024}. Moreover, the distributional shift between a fixed PRM and a continually evolving policy frequently leads to reward hacking~\citep{miao2024inform,gao2023scaling,bai2022training}. This phenomenon undermines the reliability of the policy and limits the utility of PRMs when they function as verifiers in search-based TTS.

In this paper, we investigate the effective integration of RL-based and search-based TTS for complex reasoning tasks. To eliminate the dependency on labeled process data and mitigate reward hacking risks associated with static PRMs, we propose \ours. This framework combines Adversarial Inverse Reinforcement Learning (AIRL)~\citep{airl,finn2016connection} with Group Relative Policy Optimization (GRPO)~\citep{deepseekmath2024,r1} to facilitate long CoT reasoning. During training, AIRL learns a step-wise PRM from reference trajectories. The policy model is subsequently updated using a hybrid objective that incorporates dense rewards from AIRL and binary outcome rewards from GRPO. The resulting PRM is then reused as a verifier for test-time search, creating a single pipeline in which reward learning, policy optimization, and inference-time search are trained and evaluated together.

We evaluate the policy model and PRM trained via \ours on eight standard reasoning benchmarks covering mathematics, science, and code generation. The policy model achieves an average performance improvement of $\textbf{9\%}$ over the base model, matching the performance of GPT-4o on these tasks. In search-based TTS settings, our PRM outperforms baselines trained on labeled process data across multiple policy models and datasets. We further analyze how AIRL and GRPO interact in the composite objective and how the learned PRM affects different test-time search procedures. These results suggest that \ours offers a robust and cost-efficient system for scaling test-time computation in LLMs.

Our main contributions are:
\begin{itemize}
    \item We introduce \ours, a unified framework that combines AIRL-based process reward learning, GRPO-based outcome optimization, and PRM-guided test-time search without costly step-level supervision.
    \item We provide empirical analyses of the interaction between dense process rewards and sparse outcome rewards, showing that a balanced objective improves both policy training and downstream search.
    \item Extensive evaluations on eight benchmarks demonstrate that our policy matches GPT-4o with a $\textbf{9\%}$ average improvement, while our learned PRM consistently outperforms supervised baselines across generators and search algorithms.
\end{itemize}
\vspace{-1mm}
\section{Related Works}
\vspace{-1mm}
\paragraph{Inverse Reinforcement Learning.}
Inverse reinforcement learning (IRL) \citep{abbeel2004apprenticeship,ratliff2006maximum,airl,finn2016connection} aims to recover reward functions from expert demonstrations, enabling subsequent policy training through standard RL methods. Classic IRL methods include maximum-margin IRL \citep{abbeel2004apprenticeship,ratliff2006maximum} and probabilistic maximum-entropy IRL \citep{ziebart2008maximum,ziebart2010modeling}. Adversarial IRL (AIRL) \citep{airl} reformulated IRL into a GAN-style adversarial game \citep{goodfellow2020generative}, improving generalization and theoretical grounding. Subsequent advances have further streamlined IRL: IQ-Learn optimizes inverse soft-$Q$ functions without explicitly modeling rewards \citep{garg2021iq}; Inverse Preference Learning uses offline pairwise preferences to avoid explicit reward modeling \citep{hejna2023inverse}; "IRL without RL" reframes imitation as supervised regression over trajectory data \citep{swamy2023irlfrl}. Our method adopts AIRL as a practical reward-learning module for LLM reasoning and studies how the learned step-wise reward can support both RL training and test-time search.

\paragraph{RL for LLMs.}
Reinforcement learning from human feedback (RLHF) is standard for aligning LLM outputs with user intent \citep{christiano2017deep,ouyang2022training}. Nonetheless, many open-source reasoning improvement efforts still rely on imitation learning from curated CoT datasets \citep{yuan2024reasoning,yue2024cot,wei2024cot,s1,jin2024learning,jin2025two,zhou2026dare}. Recent large-scale RL methods using sparse outcome rewards, such as OpenAI \texttt{o1} \citep{jaech2024o1} and DeepSeek-R1 \citep{r1}, have achieved significant gains. Specialized RL frameworks for mathematical reasoning, such as Math-Shepherd \citep{wang2024math} and DeepSeek-Math-7B-RL \citep{deepseekmath2024}, utilize dense supervision or extensive RL training to match larger models. A closely related work, PRIME~\citep{cui2025process}, utilizes “free process rewards”~\citep{yuan2024free} to derive an implicit token-level reward from the log-likelihood ratios between two models. While PRIME shares our motivation of leveraging implicit rewards, it differs in two ways. First, PRIME uses per-token rewards from likelihood ratios, whereas we train a step-wise discriminator—shown to provide stronger signals~\citep{nie2018relgan,rashid2024critical,stiennon2020learning}. Second, PRIME's reward is tied to the models computing the ratios, while our AIRL-based verifier transfers across generators and search algorithms.

\begin{figure*}[ht]
\vspace{-3mm}
    \centering
    \includegraphics[width=0.9\linewidth]{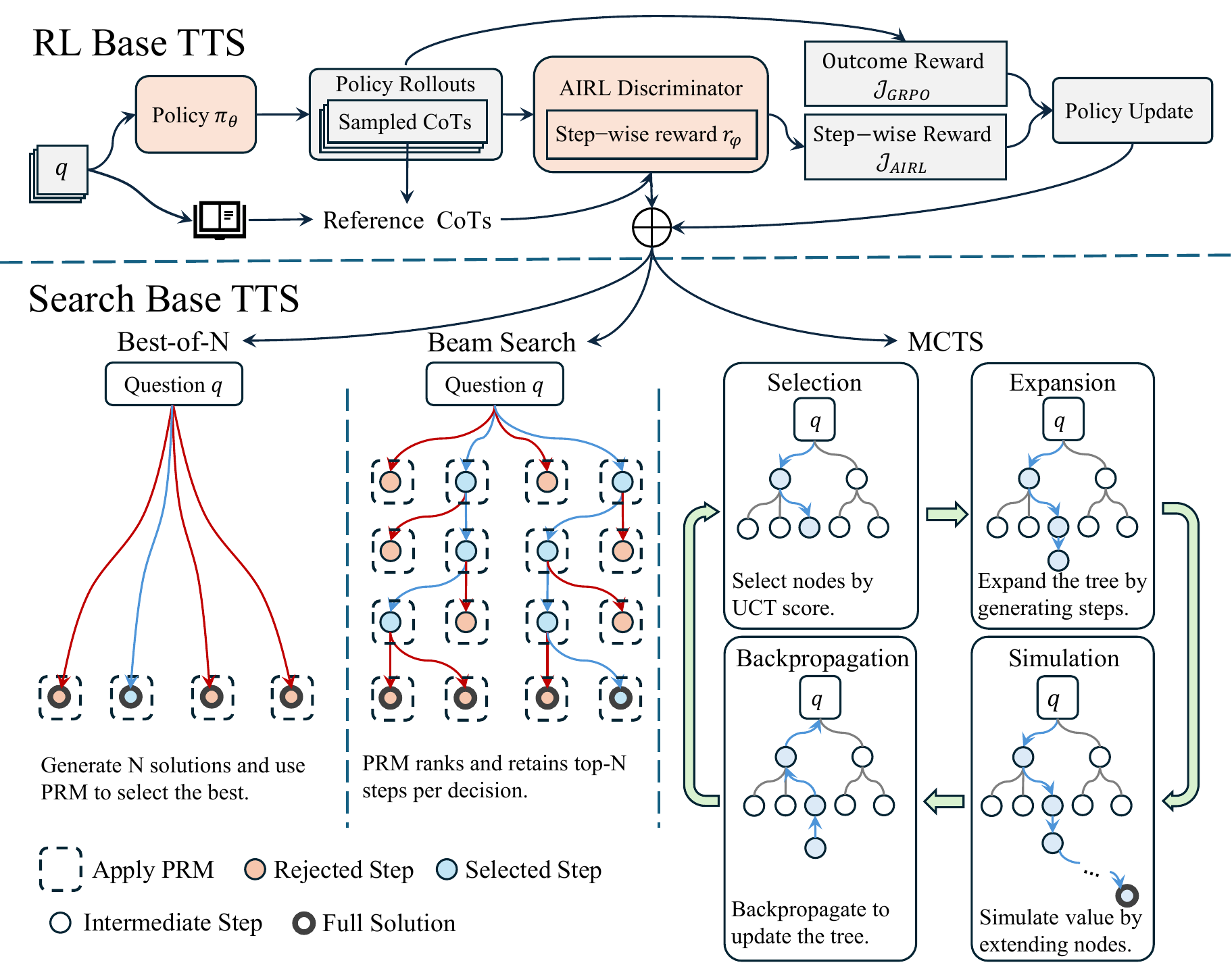}
    \caption{Overview of \ours. During training, \ours uses the AIRL discriminator to learn a PRM and optimizes the policy with both dense rewards from AIRL and outcome rewards from GRPO. At test time, the policy and PRM jointly guide downstream search.}
\label{search_method}
\vspace{-2mm}
\end{figure*}

\paragraph{Test-Time Scaling.}
TTS methods enhance the reasoning capabilities of fixed LLMs by allocating additional inference-time computation. Parallel TTS approaches aggregate independent samples to improve outcomes \citep{brown2024large,Irvine2023Rewarding,wang2023selfconsistency,zhang2023planning}. Methods such as Self-Consistency \citep{wang2023selfconsistency}, Best-of-N sampling \citep{carraro2024enhancing,snell2025scaling}, and Beam Search \citep{xie2023self,snell2025scaling} utilize diversity-driven strategies for output selection. MCTS integrates lookahead search guided by learned PRMs, achieving strong reasoning performance \citep{zhang2023planning,zhou2024language,llama_berry,xie2024mcts,park2024lemcts,rebase}. Sequential TTS refines outputs iteratively based on previous attempts \citep{s1,snell2025scaling,madaan2023self,jin2025two}. Our work integrates the AIRL-trained PRM directly into popular TTS methods, including MCTS, Beam Search, and Best-of-N, demonstrating superior performance compared to static PRMs trained on labeled data.
\vspace{-1mm}
\section{Method}
\label{sec:method}
\vspace{-1mm}

We propose \ours, a unified framework that integrates AIRL with GRPO to facilitate both effective RL training and robust test-time search. Our primary objective is to learn a generalizable, step-wise PRM without relying on expensive human annotations. This PRM provides dense supervision during training and serves as a verifier to guide search algorithms during inference. An overview of our proposed framework is presented in Figure \ref{search_method}.

\vspace{-1mm}
\subsection{Problem Setup}
\label{sec:problem_setup}
\vspace{-1mm}

Let $\mathcal{Q}$ denote a dataset of questions, where each $q \in \mathcal{Q}$ represents a specific query. A CoT is defined as a sequence of reasoning steps
\[
C(q)=\{C_1, C_2, \ldots, C_T\},
\]
that leads to a final answer. Our goal is to learn a reward function $r_{\phi}$ parameterized by $\phi$, which assigns a scalar score to each reasoning step. This function provides a dense signal to optimize the policy model $\pi_{\theta}$ via RL. During inference, the learned $r_{\phi}$ acts as a heuristic to steer the search process of $\pi_{\theta}$, thereby enhancing reasoning accuracy on unseen questions.

\vspace{-1mm}
\subsection{Data Generation and Replay Buffer}
\vspace{-1mm}
To initialize the training process, we construct a set of reference trajectories. For each question $q \in \mathcal{Q}$, we sample multiple CoTs from the current policy $\pi_{\theta}$ or utilize existing high-quality traces. We assign a binary outcome reward to each CoT based on the correctness of the final answer. Trajectories yielding a positive reward are stored in a replay buffer $\mathcal{B}$. These samples serve as the reference distribution for training the AIRL discriminator. To maintain the reliability and diversity of the reference data, we periodically prune low-quality entries from $\mathcal{B}$ and populate it with newly discovered correct solutions throughout the training process.

\vspace{-1mm}
\subsection{Learning PRM via AIRL}
\label{sec:airl}
\vspace{-1mm}
To circumvent the need for granular step-wise labels typical of static PRM training, we adopt the adversarial reward learning paradigm from AIRL~\citep{airl,goodfellow2020generative,finn2016connection,nie2018relgan}. We train a discriminator to differentiate between the reference rollouts in $\mathcal{B}$ and the samples generated by the current policy $\pi_{\theta}$. In the context of LLM reasoning, the discriminator $D_\phi$ operates on state-action pairs, defined as:
\begin{itemize}[nosep]
\item \emph{state}: the question \(q\) together with the preceding reasoning steps \(\{C_1,\dots,C_{i-1}\}\), and  
\item \emph{action}: the current step \(C_i\).
\end{itemize}
The discriminator is:
{\small
\[
D_\phi(C_i \mid q, C_{<i}) = \frac{\exp\{f_\phi(q, C_{\leq i})\}}{\exp\{f_\phi(q, C_{\leq i})\} + \pi_\theta(C_i \mid q, C_{<i})},
\]
}
where $f_\phi$ is a learned scoring function.

The intrinsic step-wise reward for a given step $C_i$ is derived as:
\begin{equation}
\label{eq:airl_reward}
r_\phi(C_i \mid q, C_{<i})=\log \frac{D_\phi(C_i \mid q, C_{<i})}
                              {1-D_\phi(C_i \mid q, C_{<i})}.
\end{equation}
And $r_\phi$ serves as the PRM.

We optimize the discriminator by minimizing the cross-entropy loss:
\begin{equation}
\label{eq:airl_loss}
\small{
\begin{split}
    \mathcal{L}_{\text{AIRL}} & = \sum_{i=1}^{|C|}\Bigl[ -\mathbb{E}_{\substack{q\sim \mathcal{Q},\\ C \sim\pi_e(\cdot\mid q)}}\!\bigl[\log D_\phi(C_i \mid q, C_{<i})\bigr] \\
    & \quad -\mathbb{E}_{\substack{q\sim \mathcal{Q},\\ C \sim\pi_\theta(\cdot\mid q)}}\!\bigl[\log\!\bigl(1-D_\phi(C_i \mid q, C_{<i})\bigr)\bigr]\Bigr],
\end{split}}
\end{equation}
where \(\pi_{e}\) denotes the reference rollout distribution drawn from the replay buffer \(\mathcal{B}\) and \(\pi_{\theta}\) is the current policy.  
Updating the discriminator can be seen as updating the reward function $r_\phi$. 

\vspace{-1mm}
\subsection{Policy Training with Combined RL Objectives}
\label{sec:grpo}
\vspace{-1mm}
The policy model is optimized using a composite objective that integrates intrinsic process rewards from AIRL and extrinsic outcome rewards via GRPO.

\paragraph{AIRL Objective.}
In the AIRL framework, the policy seeks to maximize the rewards assigned by the discriminator, effectively aligning its generation distribution with the reference distribution. We formulate the policy update using a PPO-style objective~\citep{schulman2017proximal} based on the advantage $A_i$ derived from the learned reward $r_\phi$:

{\small\begin{multline}
\label{eq:j_airl}
\mathcal{J}_{\mathrm{AIRL}}(\theta)=\mathbb{E}_{\substack{q\sim \mathcal{Q},\\ C \sim\pi_\theta(\cdot\mid q)}}\Biggl[\sum_{i=1}^{|C|} \min\biggl(\frac{\pi_\theta(C_i\mid q, C_{<i})}{\pi_{\mathrm{old}}(C_i\mid q, C_{<i})}A_i, \\
\mathrm{clip}\Bigl(\frac{\pi_\theta(C_i\mid q, C_{<i})}{\pi_{\mathrm{old}}(C_i\mid q, C_{<i})}, 1\!-\!\epsilon, 1\!+\!\epsilon\Bigr)A_i \biggr)\Biggr],
\end{multline}}
where \(C\) is a CoT produced by \(\pi_{\theta}\), \(|C|\) is its length, \(\pi_{\text{old}}\) is the sampling policy, and \(A_i\) is the advantage at step \(i\), estimated with a REINFORCE-style method~\citep{williams1992simple}.

\paragraph{GRPO Objective.}
DeepSeek-R1 achieves strong reasoning performance by pairing binary outcome rewards with GRPO~\citep{r1}. To enforce correctness on the final answer, we incorporate the GRPO objective utilized by DeepSeek-R1. For a group of $G$ rollouts $\{C^{k}\}_{k=1}^{G}$ sampled for a single question, the objective is defined as:

{\small
\begin{multline}
\label{eq:j_grpo}
\mathcal{J}_{\mathrm{GRPO}}(\theta)= \mathbb{E}_{\substack{q\sim \mathcal{Q},\{C^k\}_{k=1}^G\sim\pi_{\mathrm{old}}(\cdot\mid q)}} \Biggl[\frac{1}{G}\sum_{k=1}^G \biggl( \min\biggl(\\
\frac{\pi_\theta(C^k\mid q)}{\pi_{\mathrm{old}}(C^k\mid q)}A^k,\mathrm{clip}\Bigl(\frac{\pi_\theta(C^k\mid q)}{\pi_{\mathrm{old}}(C^k\mid q)}, 1\!-\!\epsilon,1\!+\!\epsilon\Bigr)A^k\biggr)\\
-\beta\,\mathrm{D}_{\mathrm{KL}}(\pi_\theta\Vert\pi_{\mathrm{ref}})\biggr)\Biggr],
\end{multline}}
{\small
\begin{equation}
\label{eq:kl_divergence}
\mathrm{D}_{\mathrm{KL}}\!\bigl(\pi_\theta\Vert\pi_{\mathrm{ref}}\bigr)
=\frac{\pi_{\mathrm{ref}}(C^k \mid q)}{\pi_\theta(C^k\mid q)}
-\log\!\frac{\pi_\theta(C^k\mid q)}{\pi_{\mathrm{ref}}(C^k \mid q)}
-1,
\end{equation}}

where \(\pi_{\text{ref}}\) is a frozen reference model, and \(\epsilon\) and \(\beta\) are hyper-parameters.  
The advantage \(A^{k}\) for each CoT in the group is computed from the binary outcome rewards.

\paragraph{Composite Optimization.} We define a composite objective that combines the AIRL objective and GRPO objective to incorporate both the intermediate step rewards and the outcome rewards, and update the policy model to maximize the combined RL objectives:
\begin{equation}
\label{eq:joint_obj}
{\cal J}(\theta)
=
\lambda\,{\cal J}_{\mathrm{AIRL}}(\theta)
+(1-\lambda)\,{\cal J}_{\mathrm{GRPO}}(\theta),
\end{equation}
where $\lambda$ is a hyperparameter to balance the outcome rewards and process rewards.

The training alternates between (i)~updating the discriminator $D_\phi$ to distinguish
reference rollouts from policy rollouts, and (ii)~optimizing the policy by maximizing the composite objectives of AIRL and GRPO in Equation \eqref{eq:joint_obj}. The complete training procedure is provided in Algorithm \ref{algo:airl_s} to facilitate a detailed understanding of \ours.

\vspace{-1mm}
\subsection{PRM-guided Test-Time Search}\label{sec:search}
\vspace{-1mm}
\paragraph{Test-Time Search.} During inference, search-based TTS explores multiple CoTs rather than generating a single pass.  Approaches such as Best-of-\emph{N}~\citep{snell2025scaling}, beam search~\citep{xie2023self}, and Monte Carlo Tree Search (MCTS)~\citep{zhou2024language} improve reasoning by leveraging outcome or process reward models to guide exploration.  Since $r_{\phi}$ is trained to score intermediate reasoning steps, it serves as a verifier during search. When the policy proposes several actions, either intermediate steps or complete solutions, we score each candidate with $r_{\phi}$ and retain those with the highest scores according to the chosen search strategy.  \autoref{search_method} illustrates how $r_{\phi}$ integrates with Best-of-N, beam search, and MCTS. The implementation details are provided in \autoref{sec_search_tts}.

\begin{itemize}[nosep]
\item \textbf{Best-of-\emph{N}.}  
For a given question $q$, we sample $N$ complete solutions from $\pi_{\theta}(\cdot\mid q)$. We evaluate each solution using a PRM aggregation score (defined below) and select the highest-scoring candidate.

\item \textbf{Beam Search.}  
We maintain a beam of size $N$. At each step, we expand the current nodes by generating $M$ candidate steps. These candidates are ranked by $r_{\phi}$, and the top $N$ are retained for the subsequent iteration. This process repeats until $N$ complete solutions are generated.

\item \textbf{MCTS.}  
We construct a search tree where nodes represent reasoning states. Using an expansion width $M$, we select nodes based on the Upper Confidence Bound for Trees (UCT) criterion, which balances the learned rewards from $r_{\phi}$ with visitation counts~\citep{srinivas2012information,kocsis2006bandit}. We simulate rollouts to estimate node values, back-propagate the rewards, and return $N$ solutions.
\end{itemize}

\paragraph{PRM Aggregation.}
To evaluate a full solution $C^k$ derived from search, we compute a \emph{step-wise aggregation} score based on the minimum reward along the path~\citep{snell2025scaling}:
\[
s(C^{k}) = \min_{i} r_\phi(C^{k}_{i}).
\]
To select the final answer, we employ \textbf{PRM-Min-Sum} aggregation, which sums the scores of all distinct chains leading to the same answer $a$:
\[
S(a) = \sum_{k: A^{(k)}=a} s(C^{(k)}).
\]
The answer $a$ maximizing $S(a)$ is selected as the final prediction. Further details and the definition of other PRM aggregation methods are in \autoref{sec_search_tts}.

\begin{table*}[ht]
\vspace{-3mm}
\centering
\resizebox{1\textwidth}{!}{
\renewcommand{\arraystretch}{1}
  \begin{tabular}{lccccccccc}
    \toprule
    \multirow{2}{*}[-0.8ex]{\textbf{Model}} & \multicolumn{4}{c}{\textbf{Mathematical/Scientific Reasoning}} & \multicolumn{4}{c}{\textbf{Coding}} & \multirow{2}{*}[-0.8ex]{\textbf{Avg.}} \\
    \cmidrule(lr){2-5} \cmidrule(lr){6-9}
    & \textbf{AIME2024} & \textbf{AMC} & \textbf{MATH500} & \textbf{GPQA} & \textbf{HumanEval} & \textbf{Leetcode} & \textbf{LCB-v4} & \textbf{MBPP} &  \\
    \midrule
    \multicolumn{10}{c}{\textit{API Models}} \\
    \midrule
    DeepSeek-R1
                & \textbf{79.8} & \textbf{85.5} & \textbf{96.5} & \textbf{71.2} & \textbf{97.6} & \textbf{90.4} & \textbf{71.1} & \textbf{95.8} & \textbf{86.0} \\
    DeepSeek-V3
                & 36.7 & 81.9 & 90.2 & 61.1 & 93.3 & 88.3 & 67.0 & 88.9 & 75.9 \\
    GPT-4o 
                & 13.3 & 50.6 & 65.8 & 43.9 & 90.2 & 60.6 & 44.2 & 87.3 & 57.0 \\
    \midrule
    \multicolumn{10}{c}{\textit{Local Models (RL with PRMs)}} \\
    \midrule
    Math-Shepherd-Mistral-7B-RL
                & 0.0  & 7.2  & 28.6 & 9.6  & 32.3 & 5.6  & 3.9  & 51.1 & 17.3 \\
    DeepSeekMath-7B-RL
                & 3.3  & 18.1 & 48.6 & 23.2 & 58.5 & 11.2 & 6.2  & \textbf{73.1} & 30.3 \\
    Qwen2.5-Math-7B-Instruct 
                & 13.3 & 50.6 & \textbf{79.6} & 29.3 & 57.9 & 11.7 & 9.3  & 46.0 & 37.2 \\
    Eurus-2-7B-PRIME
                & \textbf{20.0} & \textbf{56.6} & 79.2 & \textbf{33.8} & \textbf{70.7} & \textbf{31.1} & \textbf{24.3} & 70.1 & \textbf{48.2} \\
    \midrule
    \multicolumn{10}{c}{\textit{Local Models (Other Baselines)}} \\
    \midrule
    s1.1-7B
                & \textbf{16.7} & 38.6 & 72.6 & 35.4 & 76.8 & 11.1 & 9.3  & 76.7 & 42.2 \\
    Qwen2.5-7B-Instruct
                & \textbf{16.7} & 33.7 & 72.0 & 32.5 & 81.7 & \textbf{47.4} & 28.0 & \textbf{79.4} & 48.9 \\
    Phi-4-14B
                & 13.3 & \textbf{44.6} & \textbf{78.6} & \cellcolor{lightgreen}\textbf{55.6} & \textbf{84.1} & 45.6 & 31.2 & 74.3 & \textbf{53.4} \\
    \midrule
    \ourpolicy (Ours)
                & \cellcolor{lightgreen}26.7 & \cellcolor{lightgreen}59.0 & \cellcolor{lightgreen}80.2 & 40.2 & \cellcolor{lightgreen}85.1 & \cellcolor{lightgreen}54.4 & \cellcolor{lightgreen}31.3 & \cellcolor{lightgreen}83.3 & \cellcolor{lightgreen}57.5 \\
    \bottomrule
  \end{tabular}
}
\caption{Comparison of our policy model with ten SoTA open- and closed-source LLMs across eight reasoning benchmarks. \textbf{Bold} indicates the top result within each model category; \colorbox{lightgreen}{green} denotes the best result among local models. Our policy achieves the highest average performance among local models and matches GPT-4o overall. (GPQA: GPQA-Diamond; LCB-v4: LiveCodeBench-v4)}
\label{tab:main_results}
\vspace{-2mm}
\end{table*}

\vspace{-1mm}
\section{Experiments}\label{sec:eval_protocol}
\vspace{-1mm}
We conduct extensive experiments to evaluate the unified RL-based and search-based TTS framework proposed in \ours. Our evaluation focuses on four key objectives: (i) benchmarking the reasoning performance of our policy model against current state-of-the-art (SoTA) models of comparable size; (ii) comparing the effectiveness of our AIRL-derived PRM against baselines trained on labeled datasets and implicit PRMs; (iii) assessing the compatibility of our policy and PRM with various search-based TTS algorithms; and (iv) analyzing how the AIRL and GRPO objectives interact with each other and with PRM-guided search.

\vspace{-1mm}
\subsection{Experimental Details}\label{sec:experimental_details}

\paragraph{LLMs.} \textbf{Policy Baselines:} We compare our policy model against both open-source and API-based SoTA models. Open-source baselines include models optimized for reasoning via RL, such as Math-Shepherd-Mistral-7B-RL~\citep{wang2024math}, DeepSeek-Math-7B-RL~\citep{deepseekmath2024}, Qwen2.5-Math-7B-Instruct~\citep{qwen25}, and Eurus-2-7B-PRIME~\citep{cui2025process}. We also include Qwen2.5-7B-Instruct~\citep{qwen2025qwen25technicalreport} and Phi-4-14B~\citep{abdin2024phi} as general-purpose baselines. Additionally, we evaluate s1.1-7B~\citep{s1}, which is trained via direct imitation learning on DeepSeek-R1 demonstrations, to contrast with our AIRL-based approach. For API models, we select DeepSeek-R1~\citep{r1}, DeepSeekV3 (2025-03-24)~\citep{liu2024deepseekv3}, and GPT-4o (2024-11-20)~\citep{openai2024gpt4o}. \textbf{PRM Baselines:} We compare our reward model against static PRMs trained on labeled step-wise data, specifically Math-Shepherd-Mistral-7B-PRM~\citep{wang2024math} and Llama3.1-8B-PRM-DeepSeek-Data~\citep{xiong2024rlhflowmath}. We also include EurusPRM-Stage2~\citep{yuan2024free}, an implicit PRM baseline derived from an Outcome Reward Model (ORM). Further details regarding baseline specifications are provided in \autoref{sec_models_and_datasets}.

\paragraph{Tasks.} \textbf{Mathematical/Scientific Reasoning:} We validate the efficacy of \ours on AIME2024~\citep{maa2024aimei}, AMC~\citep{maa_amc}, MATH500~\citep{lightman2023let}, and GPQA-Diamond~\citep{rein2023gpqa}. \textbf{Coding:} We extend our evaluation to coding benchmarks, including HumanEval~\citep{chen2021evaluating}, MBPP~\citep{austin2021program}, LeetCode~\citep{coignion2024leetcode}, and LiveCodeBench (v4)~\citep{jain2025livecodebench}. Task specifications are detailed in \autoref{sec_models_and_datasets}.

\paragraph{Training.} Our primary training dataset consists of 160K questions, predominantly sampled from NuminaMATH~\citep{numina_math_datasets}. We utilize self-generated responses from the LLM policy as reference rollouts, generated via specific rollout prompts (see \autoref{sec_models_and_datasets}). Both the policy $\pi_\theta$ and the reward scoring function $f_\phi$ are initialized from Qwen2.5-7B-Instruct, with the head of $f_\phi$ modified for scalar-value prediction. Training is performed on 8 NVIDIA A100 GPUs using PyTorch FSDP. We employ a learning rate of $2\text{e-}6$ for the reward model and $5\text{e-}7$ for the policy model. The maximum output length is 8,192 tokens, and the global batch size is 1,024. For each question, we sample 8 outputs. The KL coefficient is set to 0.001, the weighting parameter $\lambda$ is 0.5, and training proceeds for 2 epochs.

\paragraph{Evaluation.}\label{experiments_setting} For policy-only evaluation (no search), we report zero-shot Accuracy@1 for math/science tasks and Pass@1 for coding tasks with a temperature of 0. For search-based TTS experiments, we use a temperature of 0.7 to facilitate exploration. The search tree width $M$ is set to 4, and we report the performance of the solution selected via PRM aggregation. Within each PRM comparison, we keep the generator, rollout budget, prompts, and aggregation protocol fixed to isolate the effect of the verifier. The main results are averaged over three independent runs.

\begin{figure*}[ht]
\vspace{-3mm}
    \centering
    \includegraphics[width=1\linewidth]{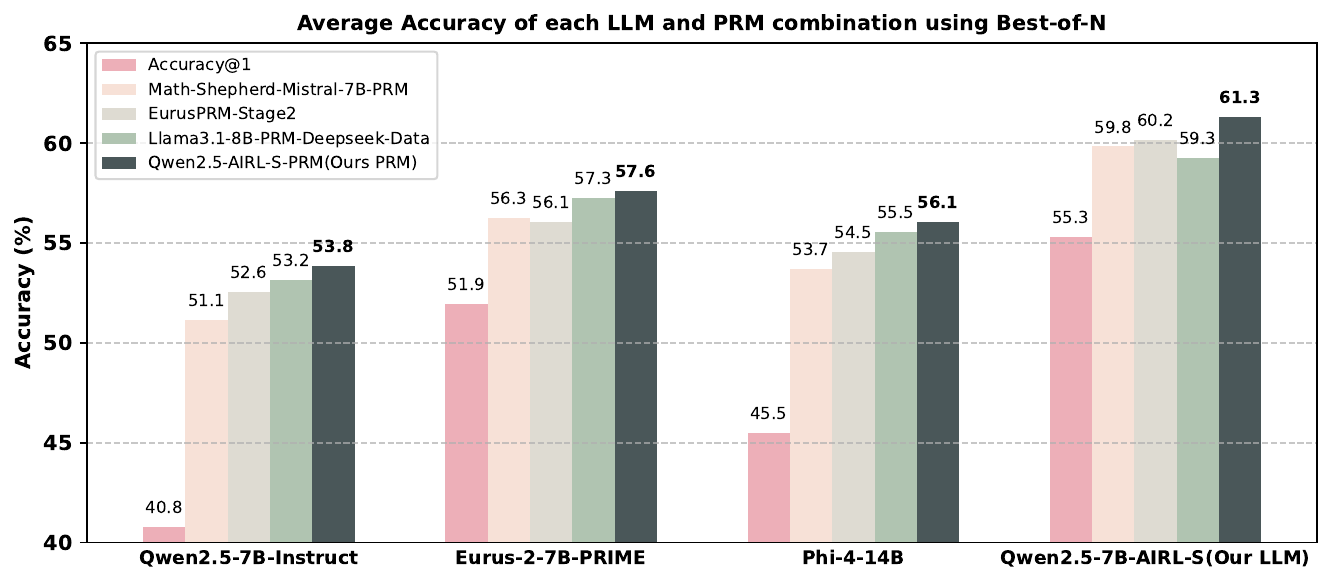}
\caption{Average performance of four PRMs applied to four generative LLMs using Best-of-N with 64 rollouts on AIME2024, AMC, and MATH500. Our PRM (Qwen2.5-AIRL-S-PRM) consistently delivers the highest test-time search performance.}
\label{fig:reward_bestofn}
\vspace{-2mm}
\end{figure*}

\begin{figure*}[t]
\vspace{-2mm}
  \centering
  \begin{subfigure}[b]{0.45\textwidth}
    \centering
    \includegraphics[width=\linewidth]{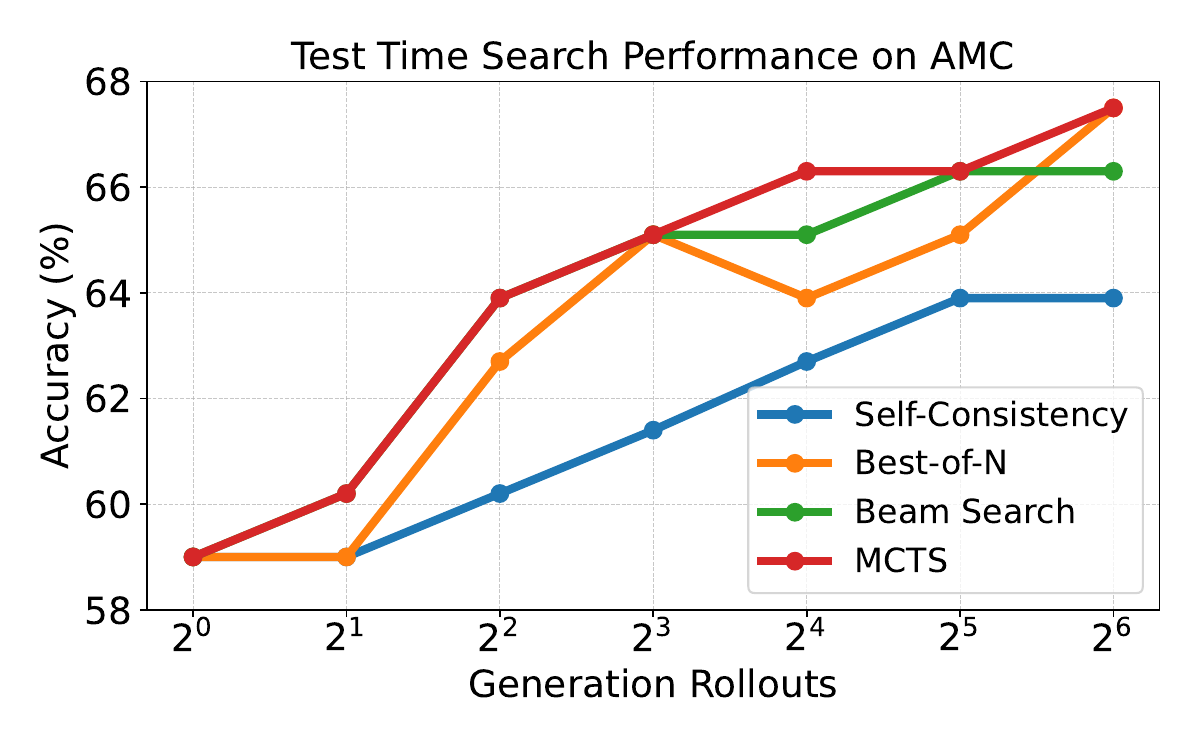}
  \end{subfigure}
  \hfill
  \begin{subfigure}[b]{0.45\textwidth}
    \centering
    \includegraphics[width=\linewidth]{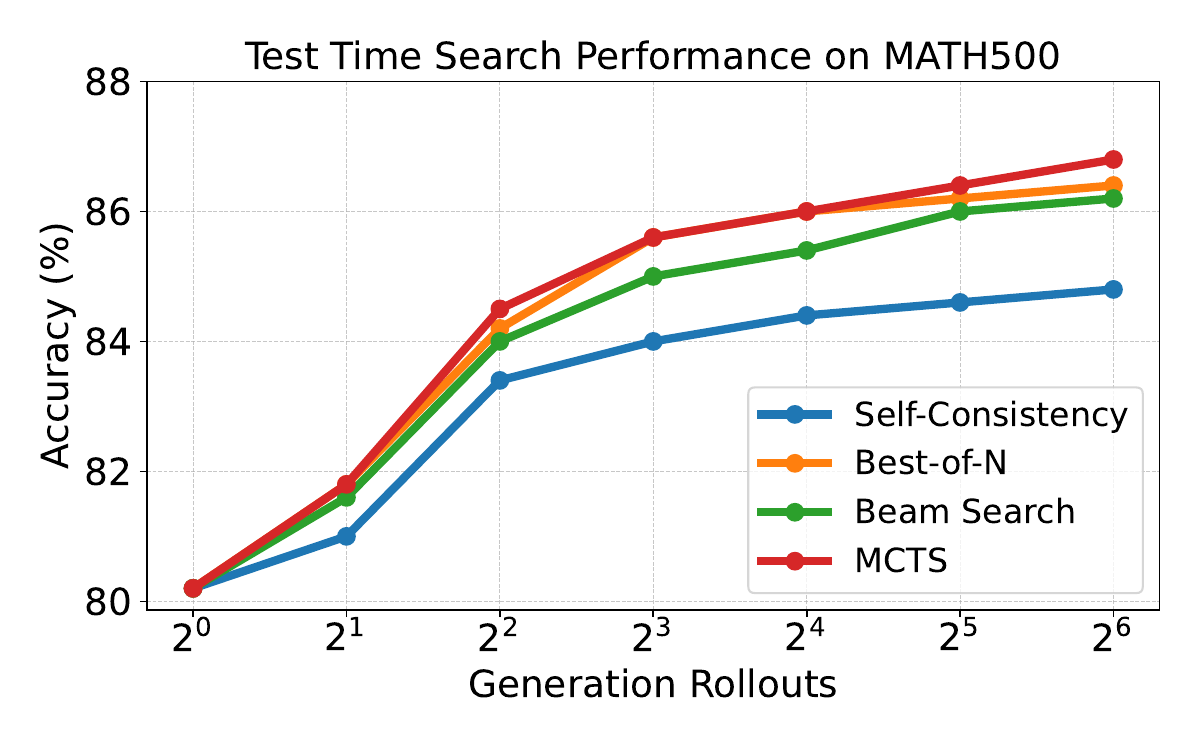}
  \end{subfigure}
  \caption{Comparison of test-time search performance with our PRM applied to MCTS, Beam Search, and Best-of-N across varying rollout counts. Our PRM consistently improves performance for all search techniques.}
  \label{fig:diff_tts}
\vspace{-2mm}
\end{figure*}

\subsection{Main Results}\label{sec:experimental_results}

\paragraph{Effectiveness of the Policy Model.} We benchmark \ourpolicy against ten SoTA models across eight reasoning tasks. As shown in \autoref{tab:main_results}, the results yield several key observations:
\ding{182} \textit{\ours significantly enhances base model performance.} When initialized with Qwen2.5-7B-Instruct, our policy achieves an average improvement of \textbf{9\%} across all benchmarks and \textbf{13\%} specifically on mathematical and scientific reasoning tasks.
\ding{183} \textit{\ours outperforms existing RL-trained baselines.} Our model surpasses Math-Shepherd-Mistral-7B-RL, DeepSeek-Math-7B-RL, and Qwen2.5-Math-7B-Instruct (all trained with explicit labeled PRMs) as well as Eurus-2-7B-PRIME (trained with implicit PRMs). This demonstrates that jointly learning step-wise rewards and optimizing the policy yields superior reasoning accuracy compared to relying on fixed supervised PRMs or implicit rewards alone.
\ding{184} \textit{Our approach outperforms direct imitation learning.} Compared to s1.1-7B, which relies on supervised fine-tuning of DeepSeek-R1 demonstrations, \ours achieves a \textbf{15\%} average improvement. This confirms the advantage of maximizing learned rewards over simple behavioral cloning.
\ding{185} \textit{Competitive efficiency.} \ours outperforms the larger Phi-4-14B on average despite possessing half the parameter count, and matches the overall performance of the general-purpose GPT-4o.

\begin{figure*}[!t]
\vspace{-3mm}
  \centering
  \begin{subfigure}[b]{0.32\textwidth}
    \centering
    \includegraphics[width=\linewidth]{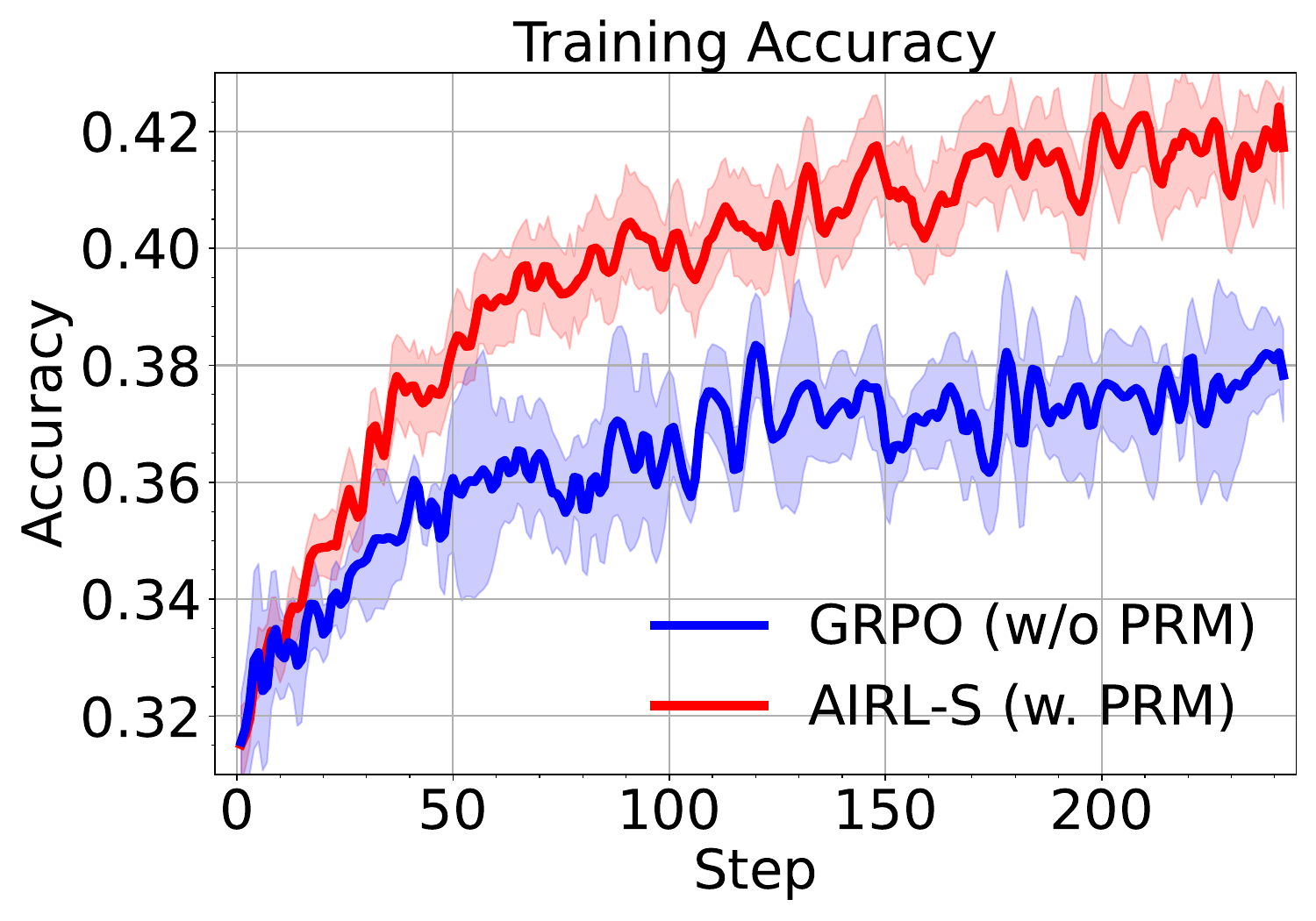}
  \end{subfigure}
  \hfill
  \begin{subfigure}[b]{0.32\textwidth}
    \centering
    \includegraphics[width=\linewidth]{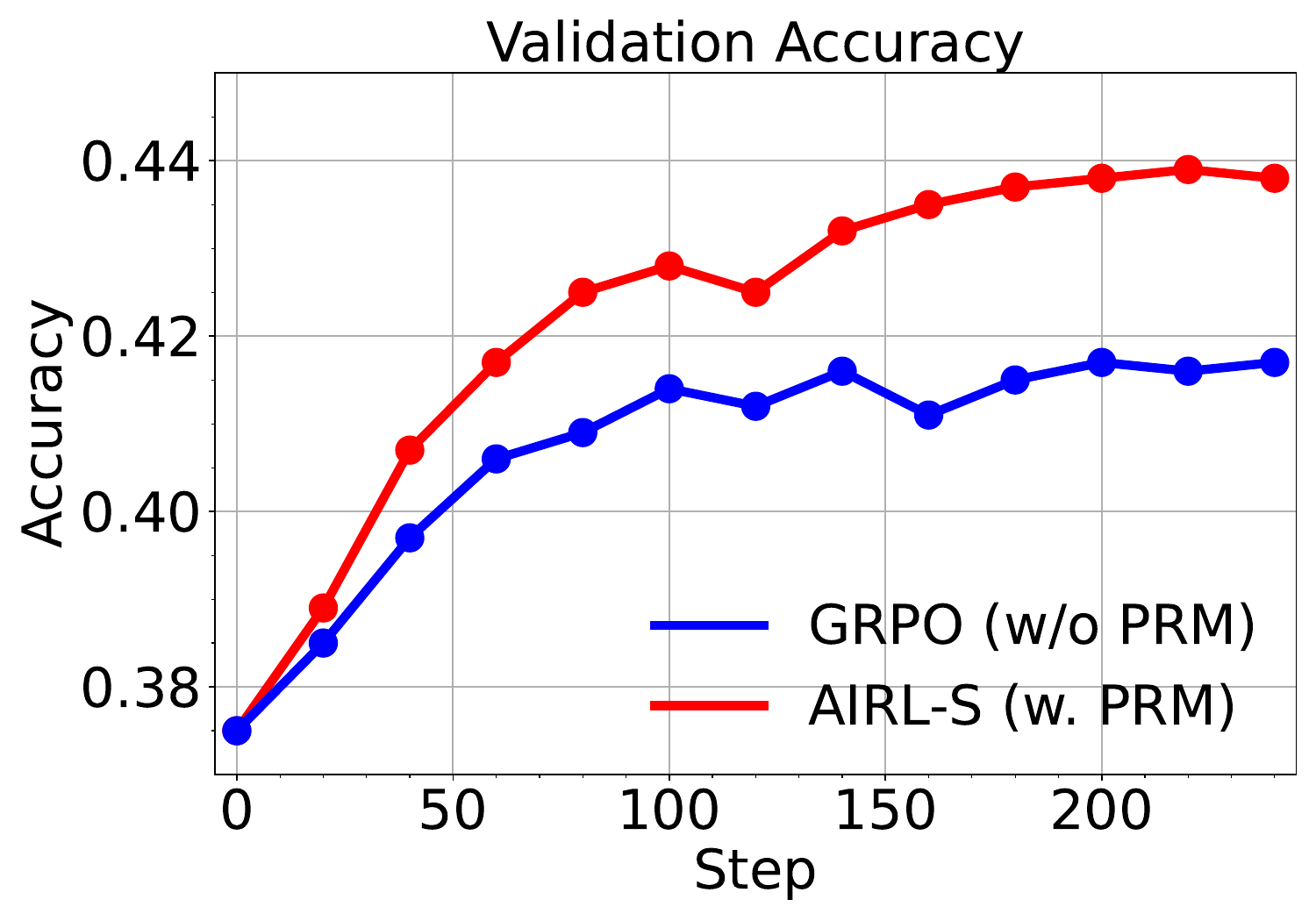}
  \end{subfigure}
    \hfill
  \begin{subfigure}[b]{0.32\textwidth}
    \centering
    \includegraphics[width=\linewidth]{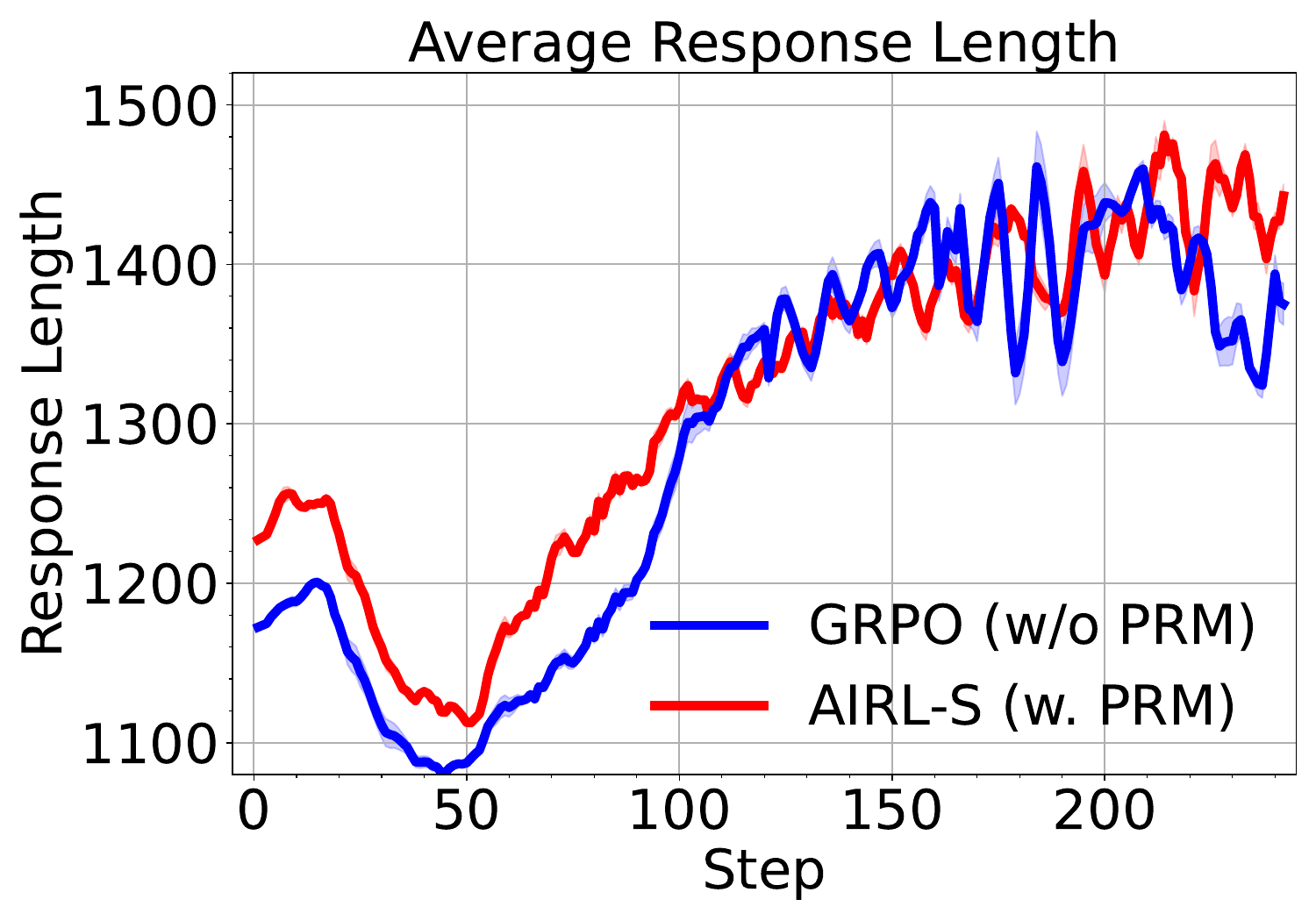}
  \end{subfigure}
  \caption{Comparison of \ours and GRPO trained with outcome rewards only. \ours improves training and validation performance and enables longer response generation at test time.}
  \label{fig:rl_training}
  \vspace{-2mm}
\end{figure*}

\begin{figure}[t]
\vspace{-3mm}
    \centering
    \includegraphics[width=\linewidth]{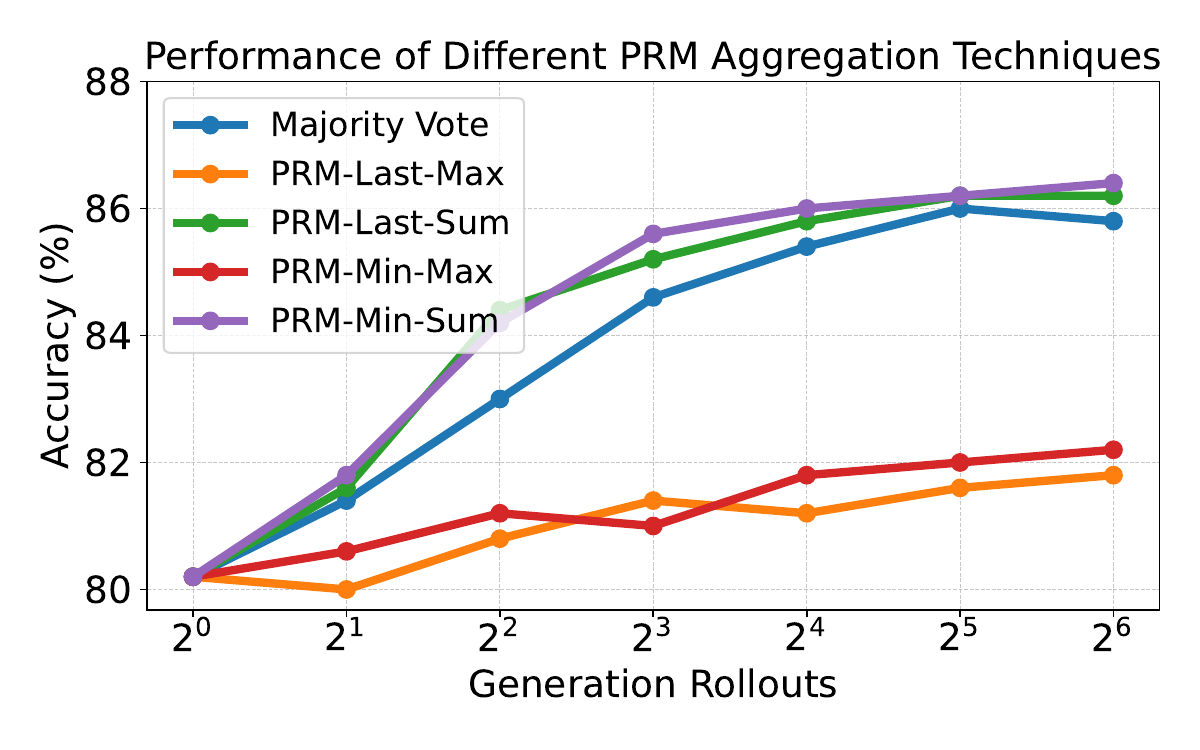}
    \caption{Performance of different PRM aggregation techniques evaluated on MATH500.}
    \label{fig:prm_aggregation_Best-of-N}
\vspace{-3mm}
\end{figure}

\paragraph{Effectiveness of the PRM.} To assess generalizability, we compare our AIRL-derived PRM (Qwen2.5-AIRL-S-PRM) against PRMs trained on labeled datasets. We conduct Best-of-N search (64 rollouts) on AIME2024, AMC, and MATH500 using four different generative LLMs. \autoref{fig:reward_bestofn} illustrates the average performance (detailed breakdowns in \autoref{sec:Evaluation Result}). We observe that:
\ding{182} \textit{Qwen2.5-AIRL-S-PRM enhances Best-of-N performance across all generator LLMs}, surpassing Math-Shepherd-Mistral-7B-PRM and EurusPRM-Stage2 by 2.4\% and 1.4\%, respectively.
\ding{183} \textit{The unified \ours framework yields optimal results.} Combining the Qwen2.5-7B-AIRL-S policy with the Qwen2.5-AIRL-S-PRM results in an \textbf{11\%} gain over the baseline Qwen2.5-7B-Instruct guided by Math-Shepherd-Mistral-7B-PRM. \autoref{tab:prm-math-shepherd} in Appendix complements this result by showing transfer to Math-Shepherd-Mistral-7B-RL, where our PRM gives the best AMC and MATH500 scores.

\paragraph{Effectiveness on Different Search-Based TTS Methods.} We integrate our PRM with MCTS, Beam Search, and Best-of-N, using Qwen2.5-7B-AIRL-S as the generator. We utilize Self-Consistency~\citep{wang2023selfconsistency} as a baseline, which relies on majority voting without reward guidance. Results on AMC and MATH500 (\autoref{fig:diff_tts}) indicate:
\ding{182} \textit{Our PRM provides robust guidance for search and selection.} Performance improves consistently across all methods as rollout counts increase, significantly outperforming Self-Consistency. This suggests that reward-guided step verification is superior to simple majority voting on final answers.
\ding{183} \textit{Value-based search methods benefit most.} The PRM is particularly effective for MCTS, where step-value estimation is critical for the UCT criterion. \autoref{tab:prm_phi4_mcts_beam} in Appendix further shows the same trend on Phi-4-14B, with our PRM obtaining the best MCTS results across all three math benchmarks.

\vspace{-1mm}
\subsection{Ablation Study}
\vspace{-1mm}

\paragraph{Impact of Our PRM on RL Training.} To isolate the contribution of the AIRL-derived reward function, we compare \ours against a baseline trained exclusively with GRPO (outcome rewards only). As illustrated in \autoref{fig:rl_training}, the integration of our PRM improves both training and validation accuracy and enables longer responses. \autoref{tab:lambda_ablation} complements this result: $\lambda=0.5$ performs best overall, while pure GRPO or pure AIRL is weaker.

\begin{table}[t]
\vspace{-1mm}
    \centering
    \resizebox{0.48\textwidth}{!}{
    \begin{tabular}{lcccc}
        \toprule
        \textbf{Method} & \textbf{AIME} & \textbf{AMC} & \textbf{MATH} & \textbf{Avg.} \\
        \midrule
        Math-Shepherd-Mistral-7B-PRM & 13.3 & 50.6 & 76.8 & 46.9 \\
        Math-Shepherd-Mistral-7B-PRM w. BoN & 20.0 & 57.8 & 82.2 & 53.3 \\
        \midrule
        Llama3.1-8B-PRM-DeepSeek-Data & 16.7 & 56.6 & 78.8 & 50.7 \\
        Llama3.1-8B-PRM-DeepSeek-Data w. BoN & 23.3 & 65.1 & 84.4 & 57.6 \\
        \midrule
        PRIME & 16.7 & 55.4 & 78.4 & 50.2 \\
        PRIME w. BoN & 20.0 & 62.7 & 84.2 & 55.6 \\
        \midrule
        \ours & 26.7 & 59.0 & 80.2 & 55.3 \\
        \ours w. BoN & \textbf{30.0} & \textbf{67.5} & \textbf{86.4} & \textbf{61.3} \\
        \bottomrule
    \end{tabular}}
    \caption{Comparison with PRIME \citep{cui2025process} and static PRM baselines under the same Qwen2.5-7B-Instruct setting. AIME, MATH, and BoN denote AIME2024, MATH500, and Best-of-N.}
    \label{tab:orm_prm_ablation}
\vspace{-3mm}
\end{table}

\paragraph{Impact of PRM Aggregation Techniques.} The method of aggregating rewards across rollouts is critical for final answer selection. We evaluate five aggregation techniques (defined in \autoref{sec_search_tts}) using Best-of-N search on MATH500. Results in \autoref{fig:prm_aggregation_Best-of-N} show that sum-based aggregation is more effective than selecting a single maximum-scoring chain, highlighting the importance of inter-answer aggregation-
weighting sum by PRM scores—over simple maximization strategies.

\paragraph{Comparison with Concurrent Implicit PRM Methods.} Finally, we compare \ours with PRIME~\citep{cui2025process}, which learns a policy-dependent reward function. Both models are trained on Qwen2.5-7B-Instruct under identical conditions. As shown in \autoref{tab:orm_prm_ablation}, \ours achieves the best average score both without search and with Best-of-N, outperforming PRIME and static PRM baselines under the same setting.

\section{Conclusion}

We present \ours, a unified framework for RL-based and search-based TTS. \ours combines AIRL and GRPO to learn a step-wise PRM from reference rollouts without labeled process data, using it for both policy training and inference-time search. Across eight benchmarks, \ours improves base model performance and outperforms RL-trained and imitation-based baselines. The learned PRM further transfers across LLMs, datasets, and search algorithms, showing that unified reward learning and test-time search provide an effective approach for complex reasoning.


\bibliography{bib/test_time_scaling,bib/other,bib/model,bib/dataset,bib/reinforcement_learning}

@article{rein2023gpqa,
  title={Gpqa: A graduate-level google-proof q\&a benchmark},
  author={Rein, David and Hou, Betty Li and Stickland, Asa Cooper and Petty, Jackson and Pang, Richard Yuanzhe and Dirani, Julien and Michael, Julian and Bowman, Samuel R},
  journal={arXiv preprint arXiv:2311.12022},
  year={2023}
}

@inproceedings{hendrycks2021measuring,
  title        = {Measuring Mathematical Problem Solving with the MATH Dataset},
  author       = {Hendrycks, Dan and Burns, Collin and Kadavath, Saurav and Arora, Akul and Basart, Steven and Tang, Eric and Song, Dawn and Steinhardt, Jacob},
  booktitle    = {Advances in Neural Information Processing Systems (NeurIPS)},
  year         = {2021},
  url          = {https://arxiv.org/abs/2103.03874}
}

@techreport{maa2024aimei,
  author      = {{Mathematical Association of America}},
  title       = {2024 American Invitational Mathematics Examination (AIME I)},
  institution = {Mathematical Association of America},
  year        = {2024},
  url         = {https://artofproblemsolving.com/wiki/index.php/2024\_AIME\_I}
}

@article{chen2021evaluating,
  title={Evaluating large language models trained on code},
  author={Chen, Mark and Tworek, Jerry and Jun, Heewoo and Yuan, Qiming and Pinto, Henrique Ponde De Oliveira and Kaplan, Jared and Edwards, Harri and Burda, Yuri and Joseph, Nicholas and Brockman, Greg and others},
  journal={arXiv preprint arXiv:2107.03374},
  year={2021}
}

@article{austin2021program,
  title={Program Synthesis with Large Language Models},
  author={Austin, Jacob and Odena, Augustus and Nye, Maxwell and Bosma, Maarten and Michalewski, Henryk and Dohan, David and Jiang, Ellen and Cai, Carrie and Terry, Michael and Le, Quoc and others},
  journal={arXiv preprint arXiv:2108.07732},
  year={2021}
}

@misc{maa_amc,
  author       = {{Mathematical Association of America}},
  title        = {{American Mathematics Competitions (AMC)}},
  howpublished = {\url{https://www.maa.org/student-programs/amc}},
  note         = {Accessed: 2025-05-05},
  year         = {2025}
}

@article{coignion2024leetcode,
  title        = {A Performance Study of LLM-Generated Code on LeetCode},
  author       = {Coignion, Tristan and Quinton, Clément and Rouvoy, Romain},
  journal      = {arXiv preprint arXiv:2407.21579},
  year         = {2024},
  url          = {https://arxiv.org/abs/2407.21579}
}

@inproceedings{
jain2025livecodebench,
title={LiveCodeBench: Holistic and Contamination Free Evaluation of Large Language Models for Code},
author={Naman Jain and King Han and Alex Gu and Wen-Ding Li and Fanjia Yan and Tianjun Zhang and Sida Wang and Armando Solar-Lezama and Koushik Sen and Ion Stoica},
booktitle={The Thirteenth International Conference on Learning Representations},
year={2025},
url={https://openreview.net/forum?id=chfJJYC3iL}
}

@article{cui2025process,
  title={Process reinforcement through implicit rewards},
  author={Cui, Ganqu and Yuan, Lifan and Wang, Zefan and Wang, Hanbin and Zhang, Yuchen and Chen, Jiacheng and Li, Wendi and He, Bingxiang and Fan, Yuchen and Yu, Tianyu and others},
  journal={arXiv preprint arXiv:2502.01456},
  year={2025}
}

@article{numina_math_datasets,
  title={Numinamath: The largest public dataset in ai4maths with 860k pairs of competition math problems and solutions},
  author={Li, Jia and Beeching, Edward and Tunstall, Lewis and Lipkin, Ben and Soletskyi, Roman and Huang, Shengyi and Rasul, Kashif and Yu, Longhui and Jiang, Albert Q and Shen, Ziju and others},
  journal={Hugging Face repository},
  volume={13},
  number={9},
  pages={9},
  year={2024}
}

@article{apps,
  author  = {Dan Hendrycks and Steven Basart and Saurav Kadavath and
             Mantas Mazeika and Akul Arora and Ethan Guo and Collin Burns and
             Samir Puranik and Horace He and Dawn Song and Jacob Steinhardt},
  title   = {Measuring Coding Challenge Competence With {APPS}},
  journal = {CoRR},
  volume  = {abs/2105.09938},
  year    = {2021},
  url     = {https://arxiv.org/abs/2105.09938},
  eprint  = {2105.09938},
  eprinttype = {arXiv}
}

@article{CodeContests,
  title   = {Competition-level code generation with AlphaCode},
  author  = {Yujia Li and David Choi and Junyoung Chung and Nate Kushman and
             Julian Schrittwieser and R{\'e}mi Leblond and Tom Eccles and
             James Keeling and Felix Gimeno and Agustin Dal Lago and Thomas Hubert
             and Peter Choy and Cyprien de Masson d’Autume and Igor Babuschkin and
             Xinyun Chen and Po-Sen Huang and Johannes Welbl and Sven Gowal and
             Alexey Cherepanov and James Molloy and Daniel J. Mankowitz and
             Esme Sutherland Robson and Pushmeet Kohli and Nando de Freitas and
             Koray Kavukcuoglu and Oriol Vinyals},
  journal = {Science},
  volume  = {378},
  number  = {6624},
  pages   = {1092--1097},
  year    = {2022},
  doi     = {10.1126/science.abq1158},
  url     = {https://www.science.org/doi/abs/10.1126/science.abq1158}
}

@article{taco,
  title   = {{TACO}: Topics in Algorithmic COde generation dataset},
  author  = {Rongao Li and Jie Fu and Bo-Wen Zhang and Tao Huang and Zhihong Sun
             and Chen Lyu and Guang Liu and Zhi Jin and Ge Li},
  journal = {arXiv preprint arXiv:2312.14852},
  year    = {2023}
}

@misc{Codeforces,
  title        = {Codeforces–Python Submissions},
  author       = {MatrixStudio},
  howpublished = {\url{https://huggingface.co/datasets/MatrixStudio/Codeforces-Python-Submissions}},
  year         = {2024}
}

@misc{s1,
  title         = {s1: Simple test-time scaling},
  author        = {Niklas Muennighoff and Zitong Yang and Weijia Shi and Xiang Lisa Li and Li Fei-Fei and Hannaneh Hajishirzi and Luke Zettlemoyer
                   and Percy Liang and Emmanuel Candès and Tatsunori Hashimoto},
  year          = {2025},
  eprint        = {2501.19393},
  archivePrefix = {arXiv},
  primaryClass  = {cs.CL},
  url           = {https://arxiv.org/abs/2501.19393}
}

@misc{omnimath,
  title         = {Omni-MATH: A Universal Olympiad Level Mathematic Benchmark For Large Language Models},
  author        = {Bofei Gao and Feifan Song and Zhe Yang and Zefan Cai and Yibo Miao and Qingxiu Dong and Lei Li and Chenghao Ma and Liang Chen
                   and Runxin Xu and Zhengyang Tang and Benyou Wang and Daoguang Zan and Shanghaoran Quan and Ge Zhang and Lei Sha and Yichang Zhang
                   and Xuancheng Ren and Tianyu Liu and Baobao Chang},
  year          = {2024},
  eprint        = {2410.07985},
  archivePrefix = {arXiv},
  primaryClass  = {cs.CL},
  url           = {https://arxiv.org/abs/2410.07985}
}

@inproceedings{lightman2023let,
  title={Let's verify step by step},
  author={Lightman, Hunter and Kosaraju, Vineet and Burda, Yuri and Edwards, Harrison and Baker, Bowen and Lee, Teddy and Leike, Jan and Schulman, John and Sutskever, Ilya and Cobbe, Karl},
  booktitle={International Conference on Learning Representations},
  volume={2024},
  pages={39578--39601},
  year={2024}
}

@article{qwen25,
  title={Qwen2. 5-coder technical report},
  author={Hui, Binyuan and Yang, Jian and Cui, Zeyu and Yang, Jiaxi and Liu, Dayiheng and Zhang, Lei and Liu, Tianyu and Zhang, Jiajun and Yu, Bowen and Lu, Keming and others},
  journal={arXiv preprint arXiv:2409.12186},
  year={2024}
}

@misc{o1,
    title = {Learning to Reason with LLMs},
    url = {https://openai.com/index/learning-to-reason-with-llms/},
    author = {OpenAI},
    month = {September},
    year = {2024}
}

@article{r1,
  title={Deepseek-r1: Incentivizing reasoning capability in llms via reinforcement learning},
  author={Guo, Daya and Yang, Dejian and Zhang, Haowei and Song, Junxiao and Zhang, Ruoyu and Xu, Runxin and Zhu, Qihao and Ma, Shirong and Wang, Peiyi and Bi, Xiao and others},
  journal={arXiv preprint arXiv:2501.12948},
  year={2025}
}

@article{liu2024deepseekv3,
  title={Deepseek-v3 technical report},
  author={Liu, Aixin and Feng, Bei and Xue, Bing and Wang, Bingxuan and Wu, Bochao and Lu, Chengda and Zhao, Chenggang and Deng, Chengqi and Zhang, Chenyu and Ruan, Chong and others},
  journal={arXiv preprint arXiv:2412.19437},
  year={2024}
}

@inproceedings{wang2024math,
  title={Math-shepherd: Verify and reinforce llms step-by-step without human annotations},
  author={Wang, Peiyi and Li, Lei and Shao, Zhihong and Xu, Runxin and Dai, Damai and Li, Yifei and Chen, Deli and Wu, Yu and Sui, Zhifang},
  booktitle={Proceedings of the 62nd Annual Meeting of the Association for Computational Linguistics (Volume 1: Long Papers)},
  pages={9426--9439},
  year={2024}
}

@misc{qwen2025qwen25technicalreport,
      title={Qwen2.5 Technical Report}, 
      author={Qwen and : and An Yang and Baosong Yang and Beichen Zhang and Binyuan Hui and Bo Zheng and Bowen Yu and Chengyuan Li and Dayiheng Liu and Fei Huang and Haoran Wei and Huan Lin and Jian Yang and Jianhong Tu and Jianwei Zhang and Jianxin Yang and Jiaxi Yang and Jingren Zhou and Junyang Lin and Kai Dang and Keming Lu and Keqin Bao and Kexin Yang and Le Yu and Mei Li and Mingfeng Xue and Pei Zhang and Qin Zhu and Rui Men and Runji Lin and Tianhao Li and Tianyi Tang and Tingyu Xia and Xingzhang Ren and Xuancheng Ren and Yang Fan and Yang Su and Yichang Zhang and Yu Wan and Yuqiong Liu and Zeyu Cui and Zhenru Zhang and Zihan Qiu},
      year={2025},
      eprint={2412.15115},
      archivePrefix={arXiv},
      primaryClass={cs.CL},
      url={https://arxiv.org/abs/2412.15115}, 
}

@techreport{openai2024gpt4o,
  author       = {{OpenAI}},
  title        = {GPT-4o System Card},
  institution  = {OpenAI},
  year         = {2024},
  month        = {November},
  day          = {20},
  url          = {https://cdn.openai.com/gpt-4o-system-card.pdf}
}

@article{abdin2024phi,
  title={Phi-4 technical report},
  author={Abdin, Marah and Aneja, Jyoti and Behl, Harkirat and Bubeck, S{\'e}bastien and Eldan, Ronen and Gunasekar, Suriya and Harrison, Michael and Hewett, Russell J and Javaheripi, Mojan and Kauffmann, Piero and others},
  journal={arXiv preprint arXiv:2412.08905},
  year={2024}
}

@article{xiong2024rlhflowmath,
  title={RLHF Workflow: From Reward Modeling to Online RLHF},
  author={Dong, Hanze and Xiong, Wei and Pang, Bo and Wang, Haoxiang and Zhao, Han and Zhou, Yingbo and Jiang, Nan and Sahoo, Doyen and Xiong, Caiming and Zhang, Tong},
  journal={arXiv preprint arXiv:2405.07863},
  year={2024}
}

@inproceedings{abbeel2004apprenticeship,
  title={Apprenticeship learning via inverse reinforcement learning},
  author={Abbeel, Pieter and Ng, Andrew Y},
  booktitle={Proceedings of the twenty-first international conference on Machine learning},
  pages={1},
  year={2004}
}

@inproceedings{ratliff2006maximum,
author = {Ratliff, Nathan D. and Bagnell, J. Andrew and Zinkevich, Martin A.},
title = {Maximum margin planning},
year = {2006},
isbn = {1595933832},
publisher = {Association for Computing Machinery},
address = {New York, NY, USA},
url = {https://doi.org/10.1145/1143844.1143936},
doi = {10.1145/1143844.1143936},
booktitle = {Proceedings of the 23rd International Conference on Machine Learning},
pages = {729–736},
numpages = {8},
location = {Pittsburgh, Pennsylvania, USA},
series = {ICML '06}
}

@inproceedings{ziebart2008maximum,
author = {Ziebart, Brian D. and Maas, Andrew and Bagnell, J. Andrew and Dey, Anind K.},
title = {Maximum entropy inverse reinforcement learning},
year = {2008},
isbn = {9781577353683},
publisher = {AAAI Press},
booktitle = {Proceedings of the 23rd National Conference on Artificial Intelligence - Volume 3},
pages = {1433–1438},
numpages = {6},
location = {Chicago, Illinois},
series = {AAAI'08}
}

@inproceedings{ziebart2010modeling,
author = {Ziebart, Brian D. and Bagnell, J. Andrew and Dey, Anind K.},
title = {Modeling interaction via the principle of maximum causal entropy},
year = {2010},
isbn = {9781605589077},
publisher = {Omnipress},
address = {Madison, WI, USA},
booktitle = {Proceedings of the 27th International Conference on International Conference on Machine Learning},
pages = {1255–1262},
numpages = {8},
location = {Haifa, Israel},
series = {ICML'10}
}

@article{garg2021iq,
  title={Iq-learn: Inverse soft-q learning for imitation},
  author={Garg, Divyansh and Chakraborty, Shuvam and Cundy, Chris and Song, Jiaming and Ermon, Stefano},
  journal={Advances in Neural Information Processing Systems},
  volume={34},
  pages={4028--4039},
  year={2021}
}

@inproceedings{hejna2023inverse,
  title={Inverse preference learning: Preference-based rl without a reward function},
  author={Hejna, Joey and Sadigh, Dorsa},
  journal={Advances in Neural Information Processing Systems},
  volume={36},
  pages={18806--18827},
  year={2023}
}

@article{swamy2023irlfrl,
  title={Inverse reinforcement learning without reinforcement learning},
  author={Swamy, Gokul and Wu, David and Choudhury, Sanjiban and Bagnell, Drew and Wu, Steven},
  booktitle={International Conference on Machine Learning},
  pages={33299--33318},
  year={2023},
  organization={PMLR}
}

@inproceedings{christiano2017deep,
  title        = {Deep Reinforcement Learning from Human Preferences},
  author       = {Paul F. Christiano and Jan Leike and Tom B. Brown and Miljan Martic and Shane Legg and Dario Amodei},
  booktitle    = {Advances in Neural Information Processing Systems 30 (NIPS 2017)},
  pages        = {4299--4307},
  year         = {2017},
  url          = {https://papers.nips.cc/paper/7017-deep-reinforcement-learning-from-human-preferences.pdf}
}

@article{ouyang2022training,
  title        = {Training Language Models to Follow Instructions with Human Feedback},
  author       = {Long Ouyang and Jeff Wu and Xu Jiang and Diogo Almeida and Carroll L. Wainwright and Pamela Mishkin and Chong Zhang and Sandhini Agarwal and Katarina Slama and Alex Ray and John Schulman and Jacob Hilton and Fraser Kelvin and Luke Miller and Maddie Simens and Amanda Askell and Peter Welinder and Paul Christiano and Jan Leike and Ryan Lowe},
  journal      = {arXiv preprint arXiv:2203.02155},
  year         = {2022},
  url          = {https://arxiv.org/abs/2203.02155}
}

@techreport{jaech2024o1,
  title        = {OpenAI o1 System Card},
  author       = {{OpenAI}},
  institution  = {OpenAI},
  year         = {2024},
  month        = {December},
  url          = {https://arxiv.org/abs/2412.16720}
}

@article{deepseekmath2024,
  title={Deepseekmath: Pushing the limits of mathematical reasoning in open language models},
  author={Shao, Zhihong and Wang, Peiyi and Zhu, Qihao and Xu, Runxin and Song, Junxiao and Bi, Xiao and Zhang, Haowei and Zhang, Mingchuan and Li, YK and Wu, Yang and others},
  journal={arXiv preprint arXiv:2402.03300},
  year={2024}
}

@article{yuan2024free,
  title={Free process rewards without process labels},
  author={Yuan, Lifan and Li, Wendi and Chen, Huayu and Cui, Ganqu and Ding, Ning and Zhang, Kaiyan and Zhou, Bowen and Liu, Zhiyuan and Peng, Hao},
  journal={arXiv preprint arXiv:2412.01981},
  year={2024}
}

@article{yuan2024reasoning,
  title   = {Agent-R: Training Language Model Agents to Reflect via Iterative Self-Training},
  author  = {Siyu Yuan and Zehui Chen and Zhiheng Xi and Junjie Ye and Zhengyin Du and Jiecao Chen},
  journal = {arXiv preprint arXiv:2501.11425},
  year    = {2025}
}

@inproceedings{yue2024cot,
  title={Mammoth: Building math generalist models through hybrid instruction tuning},
  author={Yue, Xiang and Qu, Xingwei and Zhang, Ge and Fu, Yao and Huang, Wenhao and Sun, Huan and Su, Yu and Chen, Wenhu},
  journal={arXiv preprint arXiv:2309.05653},
  year={2023}
}

@article{wei2024cot,
  title={Chain of preference optimization: Improving chain-of-thought reasoning in llms},
  author={Zhang, Xuan and Du, Chao and Pang, Tianyu and Liu, Qian and Gao, Wei and Lin, Min},
  journal={Advances in Neural Information Processing Systems},
  volume={37},
  pages={333--356},
  year={2024}
}

@inproceedings{wang2023selfconsistency,
  title={Self-consistency improves chain of thought reasoning in language models},
  author={Wang, Xuezhi and Wei, Jason and Schuurmans, Dale and Le, Quoc and Chi, Ed and Narang, Sharan and Chowdhery, Aakanksha and Zhou, Denny},
  journal={arXiv preprint arXiv:2203.11171},
  year={2022}
}

@article{xie2024mcts,
  title={Monte carlo tree search boosts reasoning via iterative preference learning},
  author={Xie, Yuxi and Goyal, Anirudh and Zheng, Wenyue and Kan, Min-Yen and Lillicrap, Timothy P and Kawaguchi, Kenji and Shieh, Michael},
  journal={arXiv preprint arXiv:2405.00451},
  year={2024}
}

@misc{park2024lemcts,
  title={Ensembling Large Language Models with Process Reward-Guided Tree Search for Better Complex Reasoning},
  author={Park, Sungjin and Liu, Xiao and Gong, Yeyun and Choi, Edward},
  journal={arXiv preprint arXiv:2412.15797},
  year={2024}
}

@article{brown2024large,
  title={Large language monkeys: Scaling inference compute with repeated sampling},
  author={Brown, Bradley and Juravsky, Jordan and Ehrlich, Ryan and Clark, Ronald and Le, Quoc V and R{\'e}, Christopher and Mirhoseini, Azalia},
  journal={arXiv preprint arXiv:2407.21787},
  year={2024}
}

@article{irvine2023rewarding,
  title={Rewarding chatbots for real-world engagement with millions of users},
  author={Irvine, Robert and Boubert, Douglas and Raina, Vyas and Liusie, Adian and Zhu, Ziyi and Mudupalli, Vineet and Korshuk, Aliaksei and Liu, Zongyi and Cremer, Fritz and Assassi, Valentin and others},
  journal={arXiv preprint arXiv:2303.06135},
  year={2023}
}

@article{carraro2024enhancing,
  title={Enhancing recommendation diversity by re-ranking with large language models},
  author={Carraro, Diego and Bridge, Derek},
  journal={ACM Transactions on Recommender Systems},
  year={2024},
  publisher={ACM New York, NY}
}

@article{goodfellow2020generative,
  title={Generative adversarial networks},
  author={Goodfellow, Ian and Pouget-Abadie, Jean and Mirza, Mehdi and Xu, Bing and Warde-Farley, David and Ozair, Sherjil and Courville, Aaron and Bengio, Yoshua},
  journal={Communications of the ACM},
  volume={63},
  number={11},
  pages={139--144},
  year={2020},
  publisher={ACM New York, NY, USA}
}

@article{srinivas2012information,
  title={Information-theoretic regret bounds for gaussian process optimization in the bandit setting},
  author={Srinivas, Niranjan and Krause, Andreas and Kakade, Sham M and Seeger, Matthias W},
  journal={IEEE transactions on information theory},
  volume={58},
  number={5},
  pages={3250--3265},
  year={2012},
  publisher={IEEE}
}

@inproceedings{kocsis2006bandit,
  title={Bandit based monte-carlo planning},
  author={Kocsis, Levente and Szepesv{\'a}ri, Csaba},
  booktitle={European conference on machine learning},
  pages={282--293},
  year={2006},
  organization={Springer}
}

@inproceedings{nie2018relgan,
  title={Relgan: Relational generative adversarial networks for text generation},
  author={Nie, Weili and Narodytska, Nina and Patel, Ankit},
  booktitle={International conference on learning representations},
  year={2018}
}

@article{rashid2024critical,
  title={A critical look at tokenwise reward-guided text generation},
  author={Rashid, Ahmad and Wu, Ruotian and Grosse, Julia and Kristiadi, Agustinus and Poupart, Pascal},
  journal={arXiv preprint arXiv:2406.07780},
  year={2024}
}

@article{stiennon2020learning,
  title={Learning to summarize with human feedback},
  author={Stiennon, Nisan and Ouyang, Long and Wu, Jeffrey and Ziegler, Daniel and Lowe, Ryan and Voss, Chelsea and Radford, Alec and Amodei, Dario and Christiano, Paul F},
  journal={Advances in neural information processing systems},
  volume={33},
  pages={3008--3021},
  year={2020}
}

@article{jin2024learning,
  title={Learning from teaching regularization: Generalizable correlations should be easy to imitate},
  author={Jin, Can and Che, Tong and Peng, Hongwu and Li, Yiyuan and Metaxas, Dimitris N and Pavone, Marco},
  journal={Advances in Neural Information Processing Systems},
  volume={37},
  pages={966--994},
  year={2024}
}

@article{zhou2026dare,
  title={DARE: Difficulty-Adaptive Reinforcement Learning with Co-Evolved Difficulty Estimation},
  author={Zhou, Yang and Jin, Can and Dong, Zihan and Wang, Zhepeng and Yang, Yanting and Zhao, Shiyu and Li, Lei and Bao, Runxue and Xie, Yaochen and Metaxas, Dimitris N},
  journal={arXiv preprint arXiv:2605.09188},
  year={2026}
}

@article{jin2025two,
  title={Two heads are better than one: Test-time scaling of multi-agent collaborative reasoning},
  author={Jin, Can and Peng, Hongwu and Zhang, Qixin and Tang, Yujin and Metaxas, Dimitris N and Che, Tong},
  journal={arXiv preprint arXiv:2504.09772},
  year={2025}
}

@article{cobbe2021training,
  title={Training verifiers to solve math word problems},
  author={Cobbe, Karl and Kosaraju, Vineet and Bavarian, Mohammad and Chen, Mark and Jun, Heewoo and Kaiser, Lukasz and Plappert, Matthias and Tworek, Jerry and Hilton, Jacob and Nakano, Reiichiro and others},
  journal={arXiv preprint arXiv:2110.14168},
  year={2021}
}

@inproceedings{li2023making,
  title={Making language models better reasoners with step-aware verifier},
  author={Li, Yifei and Lin, Zeqi and Zhang, Shizhuo and Fu, Qiang and Chen, Bei and Lou, Jian-Guang and Chen, Weizhu},
  booktitle={Proceedings of the 61st Annual Meeting of the Association for Computational Linguistics (Volume 1: Long Papers)},
  pages={5315--5333},
  year={2023}
}

@article{yu2023outcome,
  title={Outcome-supervised verifiers for planning in mathematical reasoning},
  author={Yu, Fei and Gao, Anningzhe and Wang, Benyou},
  journal={arXiv preprint arXiv:2311.09724},
  volume={2},
  number={6},
  year={2023}
}

@article{uesato2022solving,
  title={Solving math word problems with process-and outcome-based feedback},
  author={Uesato, Jonathan and Kushman, Nate and Kumar, Ramana and Song, Francis and Siegel, Noah and Wang, Lisa and Creswell, Antonia and Irving, Geoffrey and Higgins, Irina},
  journal={arXiv preprint arXiv:2211.14275},
  year={2022}
}

@inproceedings{airl,
  title={Learning Robust Rewards with Adverserial Inverse Reinforcement Learning},
  author={Fu, Justin and Luo, Katie and Levine, Sergey},
  booktitle={International Conference on Learning Representations},
  year={2018}
}

@article{finn2016connection,
  title={A connection between generative adversarial networks, inverse reinforcement learning, and energy-based models},
  author={Finn, Chelsea and Christiano, Paul and Abbeel, Pieter and Levine, Sergey},
  journal={arXiv preprint arXiv:1611.03852},
  year={2016}
}

@inproceedings{chan2024dense,
  title={Dense reward for free in reinforcement learning from human feedback},
  author={Chan, Alex J and Sun, Hao and Holt, Samuel and Van Der Schaar, Mihaela},
  booktitle={Proceedings of the 41st International Conference on Machine Learning},
  pages={6136--6154},
  year={2024}
}

@inproceedings{gao2023scaling,
  title={Scaling laws for reward model overoptimization},
  author={Gao, Leo and Schulman, John and Hilton, Jacob},
  booktitle={International Conference on Machine Learning},
  pages={10835--10866},
  year={2023},
  organization={PMLR}
}

@article{bai2022training,
  title={Training a helpful and harmless assistant with reinforcement learning from human feedback},
  author={Bai, Yuntao and Jones, Andy and Ndousse, Kamal and Askell, Amanda and Chen, Anna and DasSarma, Nova and Drain, Dawn and Fort, Stanislav and Ganguli, Deep and Henighan, Tom and others},
  journal={arXiv preprint arXiv:2204.05862},
  year={2022}
}

@inproceedings{miao2024inform,
  title={Inform: Mitigating reward hacking in rlhf via information-theoretic reward modeling},
  author={Miao, Yuchun and Zhang, Sen and Ding, Liang and Bao, Rong and Zhang, Lefei and Tao, Dacheng},
  booktitle={The Thirty-eighth Annual Conference on Neural Information Processing Systems},
  year={2024}
}

@article{schulman2017proximal,
  title={Proximal policy optimization algorithms},
  author={Schulman, John and Wolski, Filip and Dhariwal, Prafulla and Radford, Alec and Klimov, Oleg},
  journal={arXiv preprint arXiv:1707.06347},
  year={2017}
}

@article{williams1992simple,
  title={Simple statistical gradient-following algorithms for connectionist reinforcement learning},
  author={Williams, Ronald J},
  journal={Machine learning},
  volume={8},
  pages={229--256},
  year={1992},
  publisher={Springer}
}

@inproceedings{
snell2025scaling,
title={Scaling Test-Time Compute Optimally Can be More Effective than Scaling {LLM} Parameters},
author={Charlie Victor Snell and Jaehoon Lee and Kelvin Xu and Aviral Kumar},
booktitle={The Thirteenth International Conference on Learning Representations},
year={2025},
url={https://openreview.net/forum?id=4FWAwZtd2n}
}

@inproceedings{
rebase,
title={Inference Scaling Laws: An Empirical Analysis of Compute-Optimal Inference for {LLM} Problem-Solving},
author={Yangzhen Wu and Zhiqing Sun and Shanda Li and Sean Welleck and Yiming Yang},
booktitle={The Thirteenth International Conference on Learning Representations},
year={2025},
url={https://openreview.net/forum?id=VNckp7JEHn}
}

@article{zhang2023planning,
  title={Planning with large language models for code generation},
  author={Zhang, Shun and Chen, Zhenfang and Shen, Yikang and Ding, Mingyu and Tenenbaum, Joshua B and Gan, Chuang},
  journal={arXiv preprint arXiv:2303.05510},
  year={2023}
}

@article{llama_berry,
  title={Llama-berry: Pairwise optimization for o1-like olympiad-level mathematical reasoning},
  author={Zhang, Di and Wu, Jianbo and Lei, Jingdi and Che, Tong and Li, Jiatong and Xie, Tong and Huang, Xiaoshui and Zhang, Shufei and Pavone, Marco and Li, Yuqiang and others},
  journal={arXiv preprint arXiv:2410.02884},
  year={2024}
}

@article{xie2023self,
  title={Self-evaluation guided beam search for reasoning},
  author={Xie, Yuxi and Kawaguchi, Kenji and Zhao, Yiran and Zhao, James Xu and Kan, Min-Yen and He, Junxian and Xie, Michael},
  journal={Advances in Neural Information Processing Systems},
  volume={36},
  pages={41618--41650},
  year={2023}
}

@inproceedings{zhou2024language,
  title={Language agent tree search unifies reasoning, acting, and planning in language models},
  author={Zhou, Andy and Yan, Kai and Shlapentokh-Rothman, Michal and Wang, Haohan and Wang, Yu-Xiong},
  booktitle={Proceedings of the 41st International Conference on Machine Learning},
  pages={62138--62160},
  year={2024}
}

@article{madaan2023self,
  title={Self-refine: Iterative refinement with self-feedback},
  author={Madaan, Aman and Tandon, Niket and Gupta, Prakhar and Hallinan, Skyler and Gao, Luyu and Wiegreffe, Sarah and Alon, Uri and Dziri, Nouha and Prabhumoye, Shrimai and Yang, Yiming and others},
  journal={Advances in Neural Information Processing Systems},
  volume={36},
  pages={46534--46594},
  year={2023}
}

\appendix

\newpage
\section{Algorithm}
Algorithm~\ref{algo:airl_s} summarizes the \ours training procedure. All trainable components are initialized from the same pretrained checkpoint $\pi_{\theta_{\text{init}}}$, with a frozen copy $\pi_{\text{ref}}$ retained for KL regularization in the GRPO objective. The replay buffer $\mathcal{B}$ is first populated with correct rollouts from the initial policy (Line~2). Each iteration then proceeds in three phases: (i) generating $G$ candidate chains of thought per question (Lines~4--5); (ii) updating the \ours discriminator via Equation~\eqref{eq:airl_loss} to refine the step-wise reward $r_\phi$ (Line~6); and (iii) updating the policy by maximizing the composite objective in Equation~\eqref{eq:joint_obj}, which balances dense AIRL advantages with sparse GRPO outcome advantages through $\lambda$ (Line~7). The sampling policy $\pi_{\text{old}}$ is then synchronized, and newly discovered correct rollouts are added to $\mathcal{B}$ while low-quality entries are pruned (Lines~8--9). The procedure yields a jointly optimized policy $\pi_\theta$ and a transferable PRM $r_\phi$ deployable with arbitrary search algorithms at test time.

\begin{algorithm}[h]
\caption{\ours}
\label{algo:airl_s}
\textbf{Input} Initial LLM policy $\pi_{\theta_{\text{init}}}$; question set $\mathcal{Q}$; Total iterations $E$; initial replay buffer $\mathcal{B}$
\begin{algorithmic}[1] 
   \State Initialize policy model $\pi_\theta$, reference model $\pi_{\text{ref}}$, old policy $\pi_{\text{old}}$, and PRM $r_\phi$ with $\pi_{\theta_{\text{init}}}$
   \State Update replay buffer $\mathcal{B}$ by collecting correct rollouts for $q \in \mathcal{Q}$ using $\pi_\theta$
   \For{iteration = 1, \dots, $E$}
      \State Sample a batch of questions $\mathcal{B}_i$ from $\mathcal{B}$ 
      \State Generate a group of policy rollouts: $\{C^1, ..., C^G\} \sim \pi_\theta(\cdot|q)$ for $q \in \mathcal{B}_i$
      \State Update the PRM according to AIRL loss in Equation \eqref{eq:airl_loss}
      \State Update the policy $\pi_\theta$ according to the composite objectives in Equation \eqref{eq:joint_obj}
      \State Update the old policy $\pi_{\text{old}}$ using $\pi_\theta$
      \State Update the replay buffer by adding the correct rollouts to $\mathcal{B}$
   \EndFor
\end{algorithmic}
\textbf{Output} Optimized policy model $\pi_\theta$ and PRM $r_\phi$
\end{algorithm}

\section{Details for Search-based TTS}
\label{sec_search_tts}

\subsection{PRM Aggregation Techniques}

\paragraph{Answer Aggregation.} After search, we obtain $N$ candidate reasoning chains for each question $q$. Following \citet{snell2025scaling}, we compute a \emph{step-wise aggregation score} for each chain by either taking the minimum PRM reward across its steps or using the reward of its final step. For \emph{inter-answer aggregation}, we either select the answer whose chain has the highest step-wise score (Maximum Score) or sum the step-wise scores of all chains that produce the same answer and choose the answer with the highest total (Sum Score).

\paragraph{Step-wise Aggregation.} For each reasoning chain, we compute its score by one of two methods:
\begin{itemize}
    \item \textbf{Minimum Step Reward:} Use the minimum PRM reward among all steps in the chain.
    \item \textbf{Last Step Reward:} Use the PRM reward at the chain's final step.
\end{itemize}

\paragraph{Inter-answer Aggregation.} After computing step-wise scores, we aggregate across chains that yield the same final answer using one of two methods:
\begin{itemize}
    \item \textbf{Maximum Score:} Select the answer of the chain with the highest step-wise score.
    \item \textbf{Sum Score:} Sum the step-wise scores of all chains with the same answer and select the answer with the highest total.
\end{itemize}

These choices yield four PRM aggregation methods:
\begin{itemize}
    \item \textbf{PRM-Min-Max:} Minimum Step Reward + Maximum Score.
    \item \textbf{PRM-Min-Sum:} Minimum Step Reward + Sum Score.
    \item \textbf{PRM-Last-Max:} Last Step Reward + Maximum Score.
    \item \textbf{PRM-Last-Sum:} Last Step Reward + Sum Score.
\end{itemize}

In addition, we include \textbf{Majority Vote}, which selects the final answer that appears most frequently among all reasoning chains.

\subsection{Test-Time Search Techniques}

\paragraph{Best-of-N Sampling.}  
We sample \(N\) complete chains of thought (CoTs) from the policy \(\pi_\theta(\cdot\mid q)\). By default, we aggregate candidates using the PRM-Min-Sum method described in \autoref{sec:search}.

\paragraph{Beam Search.}  
We adopt the BFS-V implementation with beam size \(N\) and beam width \(M\):
\begin{enumerate}
  \item Generate \(N\) initial proposals from \(\pi_\theta\).
  \item Score each proposal using the PRM’s step-wise reward.
  \item Retain the top \(\tfrac{N}{M}\) proposals.
  \item For each surviving proposal, sample \(M\) next-step candidates and return to Step 2.
\end{enumerate}
Repeat until \(N\) complete solutions are generated. By default, we aggregate candidates with PRM-Min-Sum (\autoref{sec:search}).

\paragraph{Monte Carlo Tree Search (MCTS).}  
MCTS combines exploration via UCT with PRM-based value estimation:
\begin{enumerate}
  \item \textbf{Selection.} From the root, traverse by choosing the child \(n\) that maximizes
    \[
      \mathrm{UCT}(n) = \mu_n + \alpha \sqrt{\frac{\ln N_{\mathrm{parent}}}{N_n}},
    \]
    where \(\mu_n\) is the cumulative PRM reward of all future steps, \(N_{\mathrm{parent}}\) and \(N_n\) are the visit counts of the parent and current node, respectively, and \(\alpha=1.25\) by default.
  \item \textbf{Expansion.} At a leaf node, generate \(M\) new steps from \(\pi_\theta\).
  \item \textbf{Simulation.} Roll out each new node to completion (temperature 0) and use the terminal PRM reward as the node’s value.
  \item \textbf{Backpropagation.} Update \(\mu_n\), \(N_{\mathrm{parent}}\), and \(N_n\) along the selected path.
\end{enumerate}
Terminate when \(N\) complete solutions are obtained. By default, we use PRM-Min-Sum to aggregate candidates (\autoref{sec:search}).

\paragraph{Self-Consistency.}  
We sample \(N\) CoTs from \(\pi_\theta\) without PRM guidance and aggregate answers by majority vote (see \autoref{sec_search_tts}). Self-Consistency exploits output diversity but does not use any PRM feedback.

\section{Implementation Details}
\label{sec_models_and_datasets}

\subsection{Prompts}
\label{sec:prompts}
By default, we the same CoT prompts for both training and evaluation. The CoT prompt in our experiments are indicated in Table \ref{table_prompts}.

\begin{table*}[t]
\centering
\caption{Prompts used for training and evaluation.}
\label{table_prompts}
\renewcommand{\arraystretch}{1.4}%
\setlength{\tabcolsep}{6pt}%
\footnotesize
\begin{tabularx}{\linewidth}{@{}lX@{}}
\toprule
\textbf{Task} & \textbf{Prompt} \\ \midrule
\textsc{Mathematical/Scientific Reasoning} &
{\ttfamily
    Please reason step by step, and put your final answer within \textbackslash\textbackslash boxed\{Your answer\}.} \\[0.3em]
\textsc{Coding} &
{\ttfamily
    Write Python code to solve the problem. Please reason step by step and present the code in
    '''python
    YOUR CODE''' at the end.} \\
\bottomrule
\end{tabularx}
\end{table*}

\subsection{Models}
To clarify the context of our empirical study, \autoref{tab:BaselineModels} summarises the LLMs we utilized in our experiments, detailing their backbone architectures and key training procedures.

    \renewcommand{\tabularxcolumn}[1]{m{#1}}
    \begin{table*}[!htp]
      \centering
      \small
      \begin{tabularx}{\textwidth}{m{4cm} m{3cm} X}
        \toprule
        \textbf{Model} & \textbf{Backbone} & \textbf{Introduction} \\
        \midrule
        Eurus-2-7B-PRIME
            & Qwen2.5-Math-7B
            & Training by two stages: (i) supervised fine-tuning on the Eurus instruction corpus; (ii) online RL with PRIME implicit process reward using GSM8K, MATH, and synthetic Chain-of-Thought data \citep{cui2025process}. \\
        \midrule
        DeepSeek-Math-7B-RL
            & DeepSeek-R1-7B
            & SFT on \textit{DeepSeekMath-Instruct} (800 K math CoT pairs) followed by GRPO reinforcement learning \citep{deepseekmath2024}. \\
        \midrule
        Math-Shepherd-Mistral-7B-RL
            & Mistral-7B
            & PRM-guided step-wise PPO training on MetaMATH and MATH Level-5 problems  \citep{wang2024math}.\\
        \midrule
        Qwen2.5-Math-7B-Instruct
            & Qwen2.5-Math-7B
            & Supervised fine-tuning on 200 K math instructions plus RLHF preference optimization \citep{qwen25}. \\
        \midrule
        DeepSeek-R1
            & - 
            & Four-stage pipeline: pre-training on 1.2T tokens, SFT, RLHF, and DPO. Public endpoint \texttt{deepseek-ai/DeepSeek-R1}. We use this model for inference based on the Together API \citep{r1}. \\
        \midrule
        DeepSeek-V3
            & - 
            & Continued bilingual web-scale pre-training on DeepSeek-R1, followed by alignment stages; version tag \texttt{0324}. We use this model for inference based on the Together API  \citep{liu2024deepseekv3}.\\
        \midrule
        s1.1-7B
            & Qwen2.5-7B-Instruct
            & Few-shot SFT and low-resource RL on $\approx$1K GPT-4–generated reasoning tasks (s1K-1.1)  \citep{s1}.\\
        \midrule
        Qwen2.5-7B-Instruct
            & Qwen2.5-7B-Instruct
            & 2M general instruction pairs plus RLHF; supports 131K-token context  \citep{qwen2025qwen25technicalreport}.\\
        \midrule
        DeepSeek-R1-Distill-Qwen-7B
            & Qwen2.5-Math-7B
            & Distillation of DeepSeek-R1 pseudo-labels into Qwen-7B for reduced inference cost. \\
        \midrule
        GPT-4o
            & - 
            & OpenAI RLHF pipeline; accessed via \texttt{chat/completions} with model \texttt{gpt-4o-2024-11-20}. We use this model for inference based on the OpenAI API  \citep{openai2024gpt4o}.\\
        \bottomrule
      \end{tabularx}
            \caption{Overview of LLMs used in our experiments.}
      \label{tab:BaselineModels}
    \end{table*}

\subsection{Training Datasets}

Our training datasets consist of math questions sampled from NuminaMath \citep{numina_math_datasets}, openaimath \citep{s1}, Omni-MATH \citep{omnimath}, and code questions sampled APPS \citep{apps}, CodeContests \citep{CodeContests}, TACO \citep{taco}, and Codeforces \citep{Codeforces}. The detailed information of our training dataset is indicated in \autoref{tab:ExtraBenchmarks}.

    \renewcommand{\tabularxcolumn}[1]{m{#1}}
    \begin{table*}[!htp]
      \centering
      \small

      \begin{tabularx}{\textwidth}{m{2cm} m{2cm} X}
        \toprule
        \textbf{Dataset} & \textbf{\# Questions} & \textbf{Introduction} \\
        \midrule
        \multicolumn{3}{c}{\textit{Mathematical Reasoning}} \\
        \midrule
        NuminaMath & 125K  
            & Olympiad-level competition problems with chain-of-thought solutions; the metric is exact-match accuracy \citep{numina_math_datasets}. \\
        \midrule
        Openaimath
            & 12K & Competition-style questions released by OpenAI, spanning algebra, combinatorics, geometry, etc.; metric is exact-match accuracy \citep{s1}. \\
        \midrule
        Omni-MATH
            & 4K & Universal Olympiad-level problems covering 8 subjects and 6 difficulty tiers; evaluated via exact-match correctness \citep{omnimath}. \\
        \midrule
        \multicolumn{3}{c}{\textit{Coding}} \\
        \midrule
        APPS
            & 3K & Real-world programming challenges of varying difficulty; graded by hidden unit tests— metric is Pass@1 \citep{apps}. \\
        \midrule
        CodeContests
            & 3K & Codeforces-style contest tasks curated for AlphaCode; metric is Pass@1 \citep{CodeContests}. \\
        \midrule
        TACO
            & 10K & Algorithmic coding problems grouped by topic and difficulty; metric is Pass@1 \citep{taco}. \\
        \midrule
        Codeforces
            & 3K & A cleaned subset of Python submissions from Codeforces transformed into standalone tasks; metric is Pass@1 \citep{Codeforces}. \\
        \bottomrule
      \end{tabularx}
            \caption{Training dataset source in our experiments.}
      \label{tab:ExtraBenchmarks}
    \end{table*}

\subsection{Validation Datasets}

Our validation dataset is sampled from NuminaMath\citep{numina_math_datasets}, APPS\citep{apps}, CodeContests\citep{CodeContests}, and Codeforces\citep{Codeforces} with a total of 2K mathematical and coding questions.

\subsection{Evaluation Tasks}
For completeness, \autoref{tab:BenchmarkIntro} enumerates the evaluation tasks—covering mathematical/scientific reasoning, and code generation—on which we evaluate all systems throughout this paper.

    \renewcommand{\tabularxcolumn}[1]{m{#1}}
    \begin{table*}[!htp]
      \centering
      \small

      \begin{tabularx}{\textwidth}{l X}
        \toprule
        \textbf{Dataset} & \textbf{Introduction} \\
        \midrule
        \multicolumn{2}{c}{\textit{Mathematical/Scientific Reasoning}} \\
        \midrule
        AIME2024 
            & American Invitational Mathematics Examination 2024. Two papers (AIME I \& II) with 15 questions each, for a total of 30 short-answer problems. We evaluate \emph{exact‐match answer accuracy} and average across three random seeds for stability \citep{maa2024aimei}. \\
        \midrule
        AMC
            & All come from AMC12 2022, AMC12 2023 with 83 multiple-choice questions. Models must output the correct option; metric is accuracy \citep{maa_amc}. \\
        \midrule
        MATH500
            & A subset of the MATH benchmark containing 500 competition problems spanning seven topics and five difficulty levels. We report accuracy\citep{hendrycks2021measuring} . \\
        \midrule
        GPQA-Diamond 
            & High-quality slice of GPQA with 198 graduate-level multiple-choice questions in biology, chemistry and physics \citep{rein2023gpqa}. We report accuracy. \\
        \midrule
        \multicolumn{2}{c}{\textit{Coding Tasks}} \\
        \midrule
        HumanEval 
            & Classic Python code-generation set with 164 functions, each graded by unit tests. We report Pass@1  \citep{chen2021evaluating}. \\
        \midrule
        MBPP 
            & Sanitized version of Mostly Basic Python Problems comprising 427 hand-verified tasks; Pass@1 is reported  \citep{austin2021program}. \\
        \midrule
        LeetCode
            & 180 LeetCode tasks including 45 Easy, 91 Medium and 44 Hard; we convert to Python and grade with official test cases, reporting Pass@1 \citep{coignion2024leetcode}. \\
        \midrule
        LiveCodeBench
            & Continually-updated contest benchmark. We use releases \texttt{v4} with \textbf{776} tasks; metric is Pass@1  \citep{jain2025livecodebench}. \\
        \bottomrule
      \end{tabularx}
            \caption{Introduction of evaluation tasks used in our experiments.}
      \label{tab:BenchmarkIntro}
    \end{table*}

\section{Additional Results}
\label{sec:Evaluation Result}
\begin{figure*}[!htp]
    \centering
    \includegraphics[width=1\linewidth]{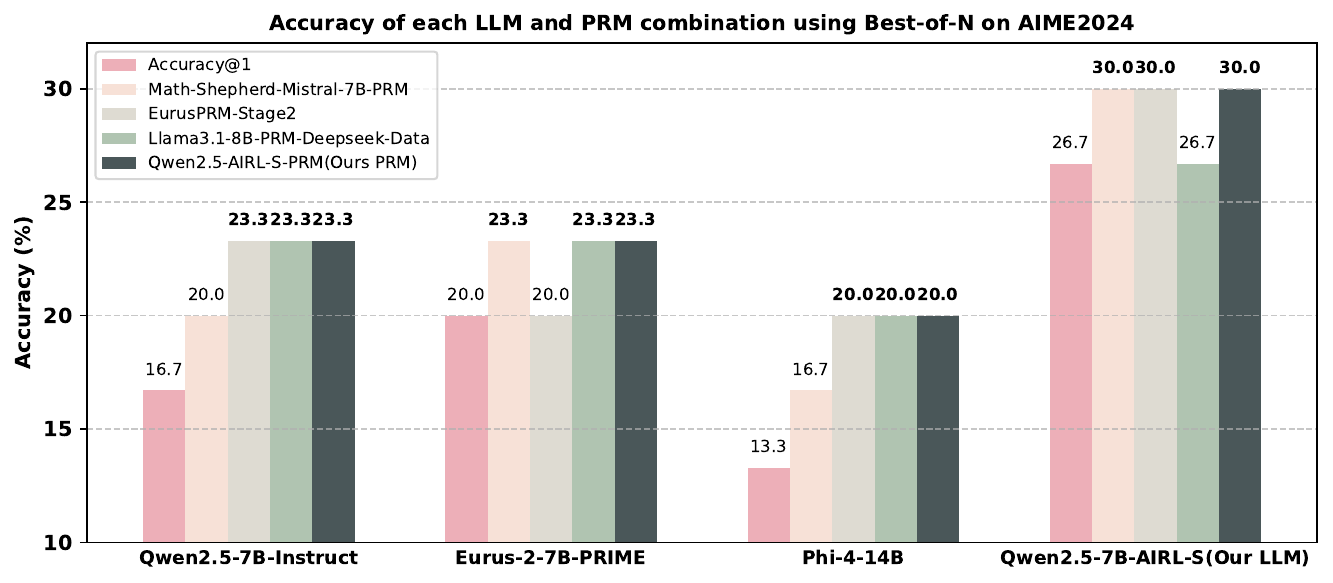}
    \caption{Performance of applying four PRMs on four different generation LLMs using Best-of-N with 64 rollouts on AIME2024}
\label{fig:reward_bestofn_aime}
\end{figure*}

\begin{figure*}[!htp]
    \centering
    \includegraphics[width=1\linewidth]{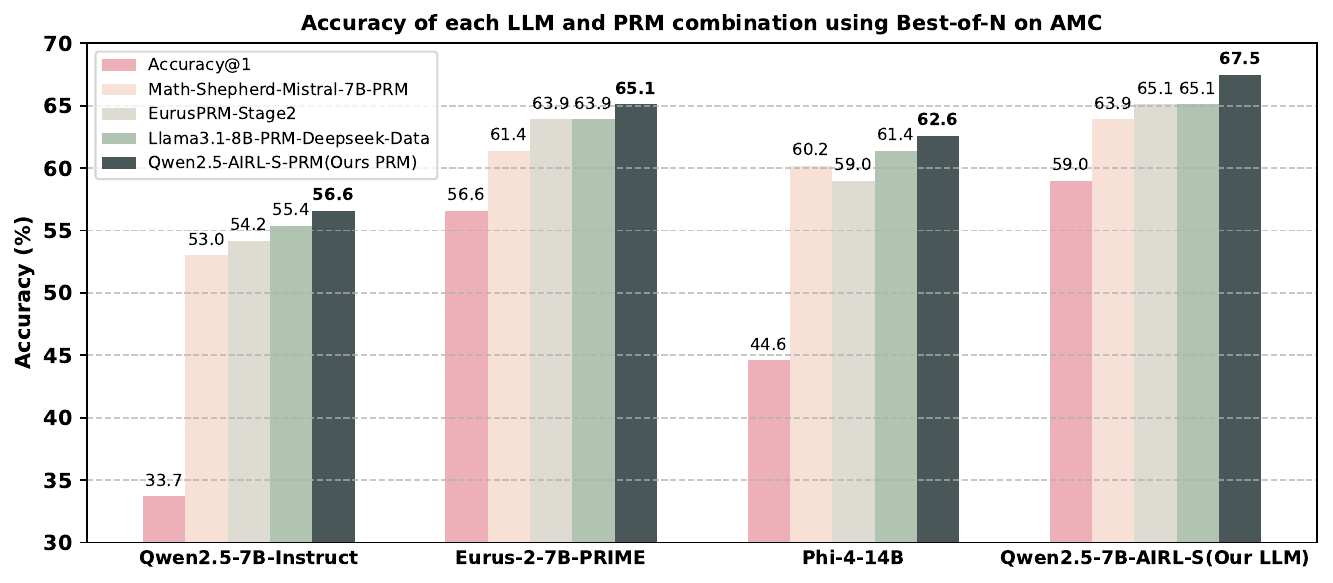}
    \caption{Performance of applying four PRMs on four different generation LLMs using Best-of-N with 64 rollouts on AMC}
\label{fig:reward_bestofn_amc}
\end{figure*}

\begin{figure*}[!htp]
    \centering
    \includegraphics[width=1\linewidth]{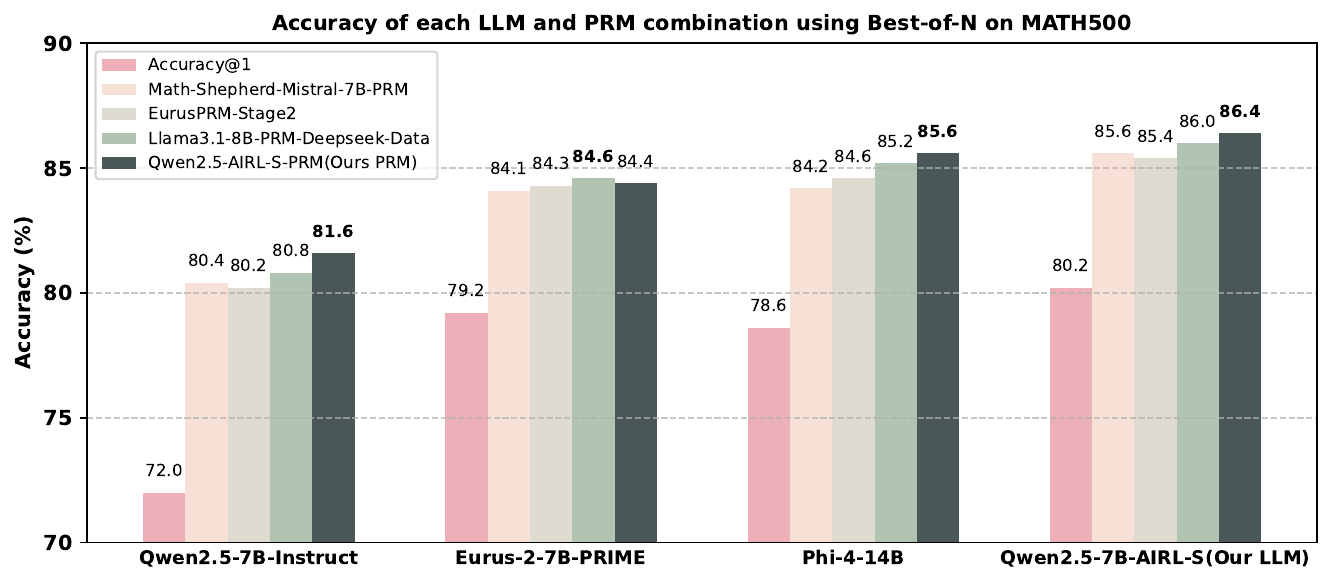}
    \caption{Performance of applying four PRMs on four different generation LLMs using Best-of-N with 64 rollouts on MATH500}
\label{fig:reward_bestofn_math}
\end{figure*}

    The detailed performance of each LLM-PRM combination using Best-of-N with 64 rollouts on AIME2024, AMC, and MATH500 are shown in \autoref{fig:reward_bestofn_aime},\autoref{fig:reward_bestofn_amc} and \autoref{fig:reward_bestofn_math}, respectively. Our policy model combined with our PRM consistently achieves the best performance in all datasets and LLM-PRM combinations, demonstrating the effectiveness of \ours in unifying RL-based and Search-based TTS.

\subsection{Additional Ablation and Component Analysis}
\label{sec:component_interaction}
We include four additional analyses to better understand how the components of \ours interact: controlled comparison against implicit PRMs, transfer to unseen generators, sensitivity to the composite objective, and compatibility with search algorithms beyond Best-of-N.

\paragraph{Controlled comparison with static and implicit PRMs.}
\autoref{tab:orm_prm_ablation} compares \ours with PRIME and two supervised static PRM baselines under the same Qwen2.5-7B-Instruct initialization. Without search, \ours reaches the highest average score (55.3), outperforming PRIME by 5.1 points and the best static PRM baseline by 4.6 points. With Best-of-N, \ours further improves to 61.3 average accuracy, exceeding PRIME by 5.7 points and the strongest static PRM baseline by 3.7 points. This controlled setting suggests that the gain does not come from a stronger backbone or a larger search budget, but from aligning the reward model with the policy distribution during training and then reusing it for search.

\paragraph{Transfer to Math-Shepherd-Mistral-7B-RL.}
\autoref{tab:prm-math-shepherd} evaluates PRMs as verifiers for Math-Shepherd-Mistral-7B-RL, whose Accuracy@1 is substantially lower than the Qwen-based policy used to train our PRM. Our PRM matches the best AIME2024 score and achieves the best AMC and MATH500 scores, improving over the no-search baseline from 7.2 to 24.1 on AMC and from 28.6 to 46.6 on MATH500. Compared with the strongest supervised PRM in this table, our verifier improves AMC by 1.1 points and MATH500 by 2.4 points, indicating that the learned reward transfers to a weaker and architecturally different generator.

\begin{table}[t]
\vspace{-1mm}
    \centering
    \resizebox{0.48\textwidth}{!}{
    \begin{tabular}{lccc}
        \toprule
        \textbf{Method} & \textbf{AIME2024} & \textbf{AMC} & \textbf{MATH500} \\
        \midrule
        Accuracy@1 & 0.0 & 7.2 & 28.6 \\
        Math-Shepherd-Mistral-7B-PRM & 3.3 & 20.7 & 42.2 \\
        EurusPRM-Stage2 & 0.0 & 21.8 & 43.4 \\
        Llama3.1-8B-PRM-DeepSeek-Data & 3.3 & 23.0 & 44.2 \\
        \textbf{Ours} & \textbf{3.3} & \textbf{24.1} & \textbf{46.6} \\
        \bottomrule
    \end{tabular}}
    \caption{Best-of-N performance of four PRMs on Math-Shepherd-Mistral-7B-RL with 64 rollouts.}
    \label{tab:prm-math-shepherd}
\vspace{-1mm}
\end{table}

\paragraph{Sensitivity to the AIRL--GRPO balance.}
\autoref{tab:lambda_ablation} studies the balance parameter $\lambda$ in the composite objective. The best overall setting is $\lambda=0.5$, which achieves the highest MATH500 and MBPP scores and ties the best AMC score. Pure GRPO ($\lambda=0$) lacks dense step-wise feedback and performs worst on MATH500 and AMC, while pure AIRL ($\lambda=1$) removes the explicit outcome signal and underperforms on AMC and MBPP. The pattern suggests a division of labor between the two objectives: GRPO anchors the model to final-answer correctness, whereas AIRL shapes intermediate reasoning steps and improves exploration. When the AIRL weight is too small, the policy receives little process-level guidance; when it is too large, the policy can over-prioritize locally plausible reasoning steps without enough pressure from final correctness. The balanced setting works best because it keeps both signals active during policy improvement.

\begin{table}[t]
\vspace{-1mm}
    \centering
    \resizebox{0.48\textwidth}{!}{
    \begin{tabular}{lccc}
        \toprule
        \textbf{$\lambda$} & \textbf{MATH500} & \textbf{AMC} & \textbf{MBPP} \\
        \midrule
        0.0 (GRPO only) & 76.6 & 54.2 & 81.6 \\
        0.25 & 79.4 & 57.8 & 82.8 \\
        \textbf{0.5} & \textbf{80.2} & \textbf{59.0} & \textbf{83.3} \\
        0.75 & 79.8 & 59.0 & 81.8 \\
        1.0 (AIRL only) & 79.4 & 57.8 & 80.7 \\
        \bottomrule
    \end{tabular}}
    \caption{Ablation of the balance parameter $\lambda$ in the composite objective.}
    \label{tab:lambda_ablation}
\vspace{-1mm}
\end{table}

\paragraph{Transfer across search algorithms and backbones.}
\autoref{tab:prm_phi4_mcts_beam} applies each PRM to Phi-4-14B with MCTS and Beam Search. Under MCTS, our PRM achieves the best result on every benchmark, improving over the strongest baseline by 3.3 points on AIME2024, 2.4 points on AMC, and 1.0 point on MATH500. Under Beam Search, our PRM obtains the best AMC score and ties the best AIME2024 and MATH500 scores. This result complements the Best-of-N experiments and shows that our PRM remains useful when the generator and search procedure both differ from those used during training. The larger gains under MCTS are consistent with how the search algorithms use PRM scores: MCTS repeatedly queries the verifier during selection, expansion, and backpropagation, so better step-value estimates compound across the tree. Beam Search also benefits from step scoring, but its narrower pruning decisions leave fewer opportunities for later PRM corrections.

\begin{table*}[t]
\vspace{-1mm}

    \centering
    \resizebox{0.9\textwidth}{!}{
    \begin{tabular}{lcccccc}
        \toprule
        & \multicolumn{3}{c}{\textbf{MCTS}} & \multicolumn{3}{c}{\textbf{Beam Search}} \\
        \cmidrule(lr){2-4} \cmidrule(lr){5-7}
        \textbf{Method} & \textbf{AIME2024} & \textbf{AMC} & \textbf{MATH500} & \textbf{AIME2024} & \textbf{AMC} & \textbf{MATH500} \\
        \midrule
        Math-Shepherd-Mistral-7B-PRM & 16.7 & 60.2 & 84.2 & 16.7 & 59.0 & 83.8 \\
        EurusPRM-Stage2 & 20.0 & 62.7 & 84.6 & 16.7 & 60.2 & 84.4 \\
        Llama3.1-8B-PRM-DeepSeek-Data & 20.0 & 62.7 & 85.2 & 20.0 & 61.4 & 84.6 \\
        Ours & \textbf{23.3} & \textbf{65.1} & \textbf{86.2} & \textbf{20.0} & \textbf{62.7} & \textbf{85.2} \\
        \bottomrule
    \end{tabular}}
    \caption{Performance of four PRMs on Phi-4-14B with MCTS and Beam Search.}
    \label{tab:prm_phi4_mcts_beam}
\vspace{-1mm}
\end{table*}

\section{Discussion}
\label{sec:appendix_discussion}

\paragraph{System-level novelty.}
\ours is designed as a unified system rather than a standalone replacement for AIRL, GRPO, or test-time search. Prior work studies these components mostly in isolation: AIRL learns rewards from demonstrations, GRPO optimizes policies with outcome rewards, and PRM-guided search improves inference by ranking candidate reasoning paths. The novelty of \ours lies in connecting these stages into one training-and-inference pipeline. The AIRL discriminator learns a step-wise PRM from reference rollouts without process labels; the policy is optimized with both dense AIRL rewards and sparse GRPO outcome rewards; and the same learned PRM is reused to guide Best-of-N, Beam Search, and MCTS at inference time.

\paragraph{Contribution beyond component reuse.}
This unification addresses two practical gaps in existing TTS systems. First, it reduces reliance on static PRMs trained from labeled process data, whose reward distribution may become mismatched with an evolving policy. Second, it links policy training and test-time search through the same verifier, allowing improvements in reward learning to benefit both stages. The results in \autoref{tab:orm_prm_ablation} show that this design outperforms static PRM and PRIME baselines under the same backbone, while \autoref{tab:prm-math-shepherd} and \autoref{tab:prm_phi4_mcts_beam} show that the learned verifier also transfers to other generators and search algorithms.

\paragraph{Interaction between objectives and search.}
The empirical behavior of \ours suggests complementary roles for its components. GRPO keeps optimization grounded in final-answer correctness, whereas AIRL supplies dense intermediate feedback for reasoning steps. As shown in \autoref{tab:lambda_ablation}, using either signal alone is weaker than balancing the two. At inference time, the PRM is most helpful when the search algorithm repeatedly uses step-level values, as in MCTS, but it also improves simpler selection methods such as Best-of-N. Thus, the main contribution is not any single component in isolation, but the way reward learning, policy optimization, and search reinforce each other within one system.

\section{Limitations}\label{sec:limitations}
Our experiments focus on reasoning tasks where final answers can be automatically verified, such as mathematics and coding. Extending \ours to open-ended generation tasks may require additional evaluation protocols. In addition, our search experiments use a fixed rollout budget and a small set of aggregation strategies; adaptive search budgets and more efficient verifier scheduling remain useful directions for future work.

\section{Impact Statement}
This work advances the reasoning capabilities of Large Language Models through efficient reinforcement learning and test-time search. By enabling smaller models to achieve high-level performance without extensive computational resources, this approach contributes to the democratization of advanced AI tools for scientific research and education. However, the amplification of reasoning power presents inherent risks, including the potential for generating sophisticated misinformation, dual-use applications in cybersecurity, and the propagation of data-driven biases during the search process.

\end{document}